\documentclass[sigconf]{acmart}

\usepackage{multirow}
\usepackage{color}
\usepackage{epsfig}
\usepackage{amsmath}

\usepackage{amssymb}
\usepackage{booktabs}
\usepackage{makecell}
\usepackage{float}
\usepackage{subcaption}
\usepackage[linesnumbered,lined,ruled]{algorithm2e}
\usepackage{algorithmic}

\usepackage{xcolor}
\usepackage{tcolorbox}
\usepackage{wrapfig}

\AtBeginDocument{%
  }

\copyrightyear{2025}
\acmYear{2025}
\setcopyright{acmlicensed}\acmConference[ICMR '25]{Proceedings of the 2025 International Conference on Multimedia Retrieval}{June 30-July 3, 2025}{Chicago, IL, USA}
\acmBooktitle{Proceedings of the 2025 International Conference on Multimedia Retrieval (ICMR '25), June 30-July 3, 2025, Chicago, IL, USA}
\acmDOI{10.1145/3731715.3733336}
\acmISBN{979-8-4007-1877-9/2025/06}

\begin{document}

\title{FedRE: Robust and Effective Federated Learning \\with Privacy Preference}

\renewcommand{\shorttitle}{Robust and Effective Federated Learning with Privacy Preference}

\author{Tianzhe Xiao}
\affiliation{%
  \institution{School of Computer Science and Technology, Huazhong University of Science and Technology}%
  \city{Wuhan}%
  \country{China}%
}
\email{d202381469@hust.edu.cn}

\author{Yichen Li}
\affiliation{%
  \institution{School of Computer Science and Technology, Huazhong University of Science and Technology}%
  \city{Wuhan}%
  \country{China}%
}
\email{ycli0204@hust.edu.cn}

\author{Yu Zhou}
\affiliation{%
  \institution{Ant Group, Chongqing Ant Consumer Finance Co., Ltd}%
  \city{Chongqing}%
  \country{China}%
}
\email{zy344525@myxiaojin.cn}

\author{Yining Qi}
\authornotemark[1]
\affiliation{%
  \institution{School of Computer Science and Technology, Huazhong University of Science and Technology}%
  \city{Wuhan}%
  \country{China}%
}
\email{qiyining@hust.edu.cn}

\author{Yi Liu}
\affiliation{%
  \institution{Ant Group, Chongqing Ant Consumer Finance Co., Ltd}%
  \city{Chongqing}%
  \country{China}%
}
\email{larry.liuy@myxiaojin.cn}

\author{Wei Wang}
\affiliation{%
  \institution{Ant Group, Chongqing Ant Consumer Finance Co., Ltd}%
  \city{Chongqing}%
  \country{China}%
}
\email{wangshi.ww@myxiaojin.cn}

\author{Haozhao Wang}
\affiliation{%
  \institution{School of Computer Science and Technology, Huazhong University of Science and Technology}%
  \city{Wuhan}%
  \country{China}%
}
\email{hz\_wang@hust.edu.cn}

\author{Yi Wang}
\affiliation{%
  \institution{Ant Group, Chongqing Ant Consumer Finance Co., Ltd}%
  \city{Chongqing}%
  \country{China}%
}
\email{haonan.wy@myxiaojin.cn}

\author{Ruixuan Li}\authornote{Corresponding author.}
\affiliation{%
  \institution{School of Computer Science and Technology, Huazhong University of Science and Technology}%
  \city{Wuhan}%
  \country{China}%
}
\email{rxli@hust.edu.cn}

\renewcommand{\shortauthors}{Tianzhe Xiao et al.}

\begin{abstract}
Despite Federated Learning (FL) employing gradient aggregation at the server for distributed training to prevent the privacy leakage of raw data, private information can still be divulged through the analysis of uploaded gradients from clients. Substantial efforts have been made to integrate local differential privacy (LDP) into the system to achieve a strict privacy guarantee. However, existing methods fail to take practical issues into account by merely perturbing each sample with the same mechanism while each client may have their own privacy preferences on privacy-sensitive information (PSI), which is not uniformly distributed across the raw data. In such a case, excessive privacy protection from private-insensitive information can additionally introduce unnecessary noise, which may degrade the model performance. In this work, we study the PSI within data and develop \textbf{\textit{FedRE}}, that can simultaneously achieve robustness and effectiveness benefits with LDP protection. More specifically, we first define PSI with regard to the privacy preferences of each client. Then, we optimize the LDP by allocating less privacy budget to gradients with higher PSI in a layer-wise manner, thus providing a stricter privacy guarantee for PSI. Furthermore, to mitigate the performance degradation caused by LDP, we design a parameter aggregation mechanism based on the distribution of the perturbed information. We conducted experiments with text tamper detection on T-SROIE and DocTamper datasets, and FedRE achieves competitive performance compared to state-of-the-art methods.
\end{abstract}

\ccsdesc[300]{Security and privacy~Distributed systems security}

\keywords{Federated Learning, Local Differential Privacy, Privacy Preference, Privacy-Sensitive Information}

\maketitle

\section{Introduction}
\label{sec:intro}
Federated Learning (FL) has emerged as a basic paradigm that enables multiple parties to jointly train a model through the aggregation of parameters without sharing their private dataset~\cite{Mcmahan17, Liu22}. Due to the benefits of preserving privacy and communication efficiency, FL has been widely deployed in various applications, such as smart healthcare ~\cite{Antunes22,Rahman2023} and finance analysis~\cite{Yang19,Long2020,byrd2020}.

However, FL is not always impervious to security threats. A notable threat namely \textit{gradient leakage attack}, aims to infer sensitive information from the shared model updates (gradients)~\cite{Li23}. Then, the malicious participant can exploit this information to reconstruct the private data or infer properties about it, leading to the violation of local privacy~\cite{Zhu19, Geiping20}.

\begin{figure}[t]
\centering
\includegraphics[width=\linewidth]{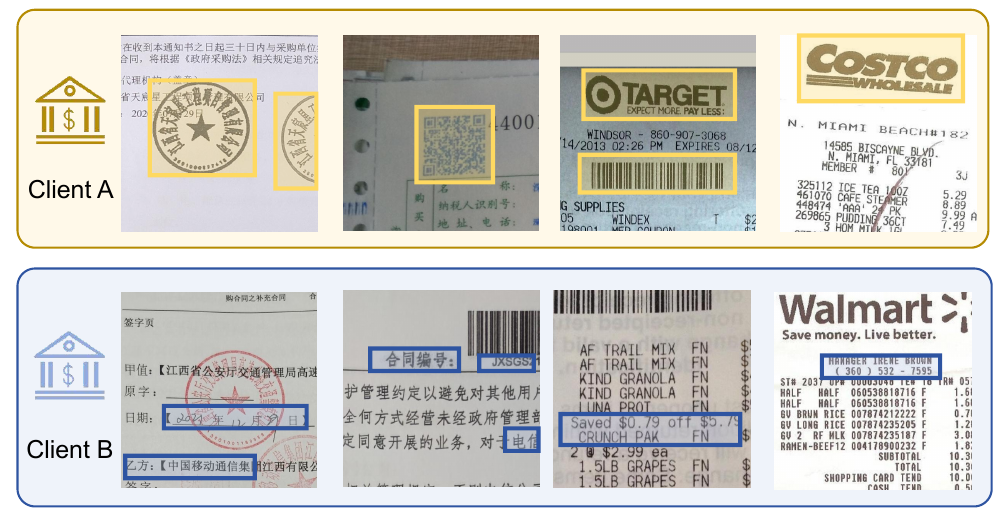}
\caption{This illustration depicts the varied privacy preferences of different clients. Client A pays close attention to iconographic elements within the data, such as stamps, bar codes, and QR codes, and highlights these privacy-sensitive regions with a yellow border. In contrast, Client B is more concerned with textual content, including numerical values and phone numbers, marking these sections with a blue frame.}
\label{fig1}
\end{figure}

\begin{figure}[t]
\centering
\resizebox{0.7\linewidth}{!}{\includegraphics{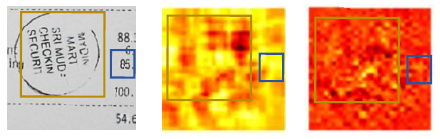}}
\caption{Different derivatives of different layer networks on input images (left : original image, middle and right : derivatives on different layers). }
\label{fig6}
\end{figure}

To address this issue, several defense methods have been proposed to mitigate the risk of gradient leakage attacks in FL~\cite{Wang22, Huang21}. LDP-Fed ~\cite{Truex20} is proposed to optimize LDP for the FL system which ensures a lightweight and quantifiable measure for privacy preservation. The authors in ~\cite{cheng2022} aim to solve performance degradation in FL with user-level DP and employ regularization and sparsification techniques to local updates. FedDPA is proposed in ~\cite{yang2023} to study the differential privacy in the personalized FL with dynamic fisher personalization and adaptive constraint. PrivateRec ~\cite{liu2023} focuses on federated recommendation scenarios and is devoted to achieving better utility in online serving under a DP guarantee.

While these approaches have achieved great success, they are
generally designed for the scenario where the same privacy protection mechanism is applied to all samples, overlooking two critical realities: First, clients are likely to have distinct privacy preferences, meaning varied privacy-sensitive information (PSI) for protection. The complexity of this issue arises from the attributes of privacy preference, which are not always explicitly observed in the data. PSI of privacy preferences may not be as overt as specific pixel regions within an image or particular words within a text. Rather, they may hinge on the overarching structure of the data, encompassing semantic information. Second, PSI is not uniformly distributed across data. Implementing a sample and indiscriminate privacy protection approach across all data samples can lead to the introduction of unnecessary noise, which can adversely affect model performance.

Considering that we are the first to propose and explore the protection of privacy preferences of clients in FL, in this paper, we seek to explore a foundational privacy protection scenario, namely \textit{privacy-sensitive regions} in images. We illustrate this concept in Fig. \ref{fig1}. To explore this scenario, we have devised a robust and effective strategy for the preservation of PSI in regions. Local Differential Privacy~\cite{arachchige2019local,cormode2018privacy} have demonstrated great success against gradient leakage attacks by perturbing samples with the privacy budget. A direct idea to solve the problem is to apply these LDP-based methods to perturb the designated privacy-sensitive regions with a lower privacy budget, which provides a stricter privacy guarantee. Despite the simplicity of such an approach, applying direct pixel-level perturbation results in the loss of critical feature information, which in turn compromises model performance~\cite{zhao2019differential}. This presents a challenge in striking the optimal balance between the robustness of the privacy protection and the effectiveness of the model accuracy.

To tackle this challenge, we propose FedRE - which can ensure both robustness and effectiveness benefits in the FL system. Figure \ref{fig6} shows the derivatives calculated by different network layers on the same input image. We observe that if the sensitive regions selected by the client are different (yellow or blue boxes), the derivative values accumulated by different regions in different layers of the network are different. Inspired by ~\cite{Wang22}, we study the layer-wise information leakage from the gradients, using the sensitivity of gradient changes regarding the PSI region to quantify the leakage risk. Then, we allocate different privacy budgets to perturb the gradients of each layer guided by the sensitivity. 
To mitigate the adverse effects that local gradient perturbation may have on the performance of the global model, we introduce a new aggregation mechanism. Upon receiving gradients from local clients, the server employs a publicly available dataset to evaluate the sensitivity of these gradients. The global model will favor aggregating less sensitive local gradients, which can reduce the infusion of noise from the local perturbed gradients, thereby preserving the effectiveness of the global model.

To verify the effectiveness of our method, we first manually annotate the PSI region of two real-world datasets: T-SROIE and DocTamper. Based on these datasets, extensive experiments have been done and show that the proposed FedRE enables more accurate and robust models relative to state-of-the-art baselines. The major contributions of this paper are summarized as follows:
\begin{itemize}
\item We propose and formally define the concept of privacy preferences in the context of federated learning, highlighting the need to protect privacy-sensitive information (PSI) in privacy-sensitive regions. Our definition accounts for the diverse and unique privacy concerns of different clients, acknowledging that PSI can vary significantly between data and clients.
\item We introduce \textbf{FedRE}, a novel method that integrates local differential privacy (LDP) in a layer-wise manner to provide tailored privacy protection for PSI. Our approach judiciously allocates the privacy budget across layers based on the sensitivity of the gradients to PSI, allowing for a more nuanced and effective privacy guarantee without substantially compromising on the model's performance.
\item We conduct extensive experiments on the annotated PSI regions of the T-SROIE and DocTamper datasets to validate the effectiveness of our proposed method. Our empirical results demonstrate that FedRE achieves superior performance in terms of robustness and accuracy compared to existing state-of-the-art methods, thereby confirming the practical utility of our approach in real-world FL scenarios.
\end{itemize}

\begin{figure*}[t]
\centering
\includegraphics[width=\linewidth]{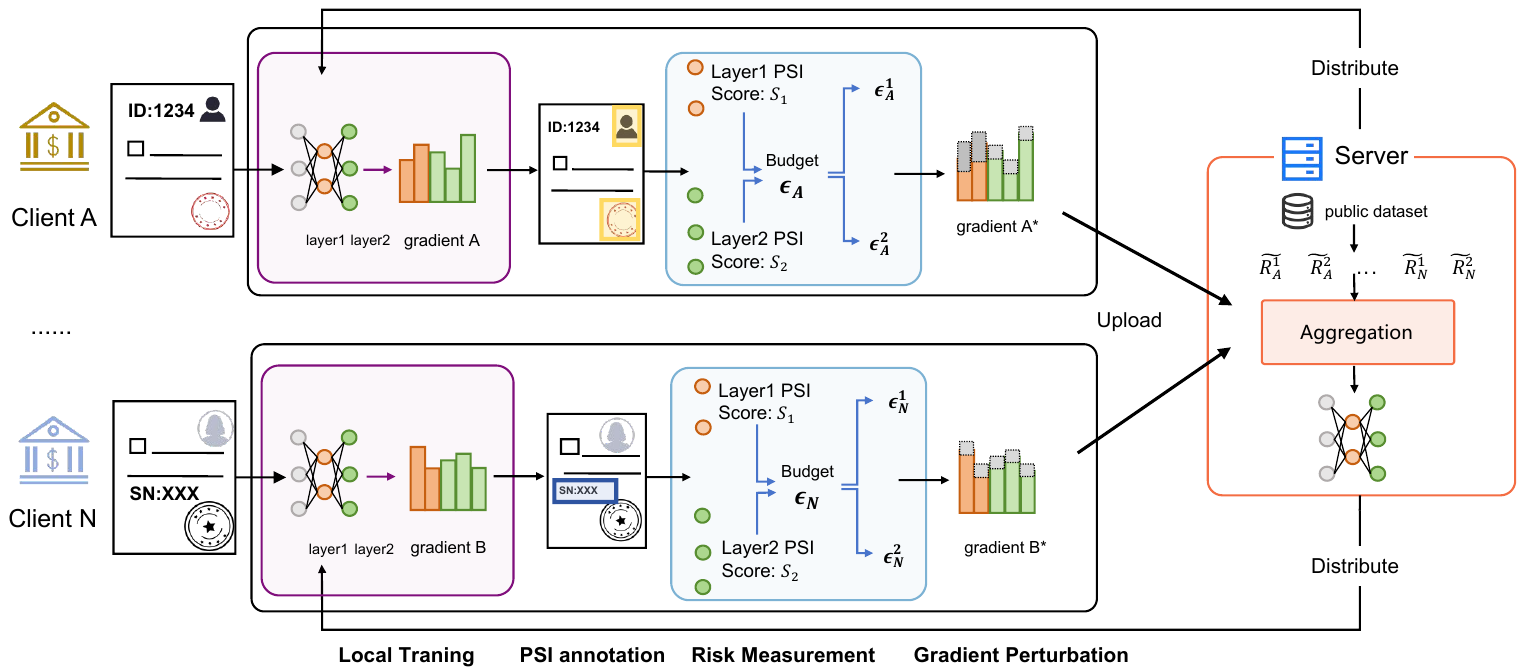}
\caption{The overall architecture of FedRE. When clients want to train a model cooperatively, each of them first trains locally on the last round's global model to get the original gradient. Then based on annotated PSI, we can measure the risk of PSI leakage at each layer by calculating their PSI scores, and the privacy budget $\epsilon$ will be rationally allocated to each layer accordingly. The original gradient will then be perturbed according to the privacy budget of each layer correspondingly and uploaded to the server for aggregation. Finally, in order to reduce the degradation caused by perturbation, the server aggregates all the gradients based on the possible distribution of the perturbation information known using the public dataset and distributes the updated model to all clients for the training of the next round.}
\label{framework}
\end{figure*}

\section{Related Work}
\label{sec:related}

\noindent\textbf{Federated Learning.}
Federated Learning is a distributed machine learning approach that enables multiple entities to collaboratively train a model without directly sharing raw data, thereby preserving data privacy and security~\cite{kairouz2021advances,li2020federated}. Key research directions in this field encompass algorithm optimization for efficient learning~\cite{wang2023dafkd,wu2023serverless}, security and privacy enhancements~\cite{qin2023revisiting,yi2023ua}, system and architectural design for scalability~\cite{qin2023fedapen}, graph learning ~\cite{gu2023dynamic,pan2023lumos},continual learning~\cite{luopan2023fedknow,li2024towards}, and incentive mechanisms to encourage and select participation~\cite{sun2023shapleyfl,yan2023criticalfl}.

Large corporations have shown significant interest in federated learning for various applications. For instance, in the financial sector, banks can leverage federated learning for enhanced fraud detection~\cite{yang2019ffd,zheng2021federated}, allowing them to share insights from their models without revealing sensitive transaction data. Other applications include personalized recommendations in e-commerce and news ~\cite{liu2023,yi2023ua}, medical research in healthcare~\cite{rieke2020future,Rahman2023}, and device optimization in manufacturing~\cite{khan2020resource,deng2023federated}. This approach enables the collective enhancement of various capabilities while maintaining strict data privacy.

\noindent\textbf{Gradient leakage attack.}
Federated learning is susceptible to malicious attacks, including gradient leakage~\cite{wei2020framework}, model inversion~\cite{li2022ressfl}, and membership inference~\cite{zhang2020gan} attacks. Gradient leakage is particularly harmful as it can reveal extensive information from the victim's training data. This attack method involves initializing pseudo training data and labels, and optimizing them to mirror real gradients. As the pseudo and real gradients converge, the pseudo data begins to reflect the properties of the actual private data. 

The method proposed in ~\cite{Zhu19} can effectively attack not only computer vision tasks but also natural language processing tasks. Subsequent work has improved in areas such as initialization with prior knowledge~\cite{Jeon21}, ground-truth label extraction~\cite{Zhao20}, faster optimizer~\cite{Geiping20}, and regularization terms~\cite{Geiping20,Yin21}. These improvements enable more effective gradient leakage attacks on larger batch sizes, higher resolutions, and more complex models (such as ViT)~\cite{Yin21}. Therefore, understanding and mitigating gradient leakage attacks is crucial due to their potential to cause significant harm.

\noindent \textbf{Privacy Protection.}
Privacy protection in machine learning encompasses various techniques aimed at safeguarding sensitive information. Differential privacy is a technique that introduces noise to the gradients before they are shared, thereby limiting the amount of information that can be inferred from them~\cite{Truex20,Wei20}. This method provides a mathematical guarantee of privacy but at the cost of model accuracy. Secure aggregation is another technique where the gradients are encrypted in a way that allows the server to compute their sum without being able to decrypt individual gradients~\cite{Bonawitz17}. This method provides robust security guarantees but requires more computational resources. Homomorphic encryption is a cryptographic technique that allows computations to be performed on encrypted data without decrypting it, providing another layer of security~\cite{Zhang20}.

However, a uniform privacy protection mechanism based on these techniques is deployed across all samples, ignoring the PSI distribution within data and the privacy preference.

\section{Methodology}
We first formulate privacy preference scenarios and propose the robust and effective FedRE. Then, we present a scalable algorithm and provide rigorous analytical results to show the efficiency of the proposed method.
\subsection{Problem Formulation}
\noindent\textbf{FL Procedures.} 
Our work is developed based on the paradigm of Federated Averaging (\texttt{FedAvg}) algorithm.  \texttt{FedAvg}, introduced by Google in 2016 ~\cite{Mcmahan17}, is a seminal work in the domain of federated learning. Initially designed for privacy-preserving machine learning on mobile devices, \texttt{FedAvg} transcends mobile applications to enable collaborative model training across multiple institutions. In such federated settings, institutions maintain the privacy of their local data while collectively training a global model through the exchange of model updates, not raw data.

Based on \texttt{FedAvg}, we aim to collaboratively train a global model for $K$ total clients in FL. We consider each client $k$ can only access to his local private dataset $D_k:=\{x_i, y_i\}$, where $x_i$ is the $i$-th input data sample and $y_i \in \{1,2,\cdots,C\}$ is the corresponding label of $x_i$ with $C$ classes. The global dataset is considered as the composition of all local datasets $D=\sum_{k=1}^{K}D_k$. The objective of the FL learning system is to learn a global model $w$ that minimizes the total empirical loss over the entire dataset $D$:
\vspace{-5pt}
\begin{align}
    &\min_{w} \mathcal{L}(w):= \sum_{k=1}^K \frac{|D_k|}{|D|}\mathcal{L}_k(w), \nonumber \\ 
    &\text{where} \ \mathcal{L}_k(w) = \frac{1}{|D_k|} \sum_{i=1}^{|D_k|} \mathcal{L}_{CE}(w; x_i, y_i),
    \label{fl_loss}
\end{align}
where $\mathcal{L}_k(w)$ is the local loss in the $k$-th client and $\mathcal{L}_{CE}$ is the cross-entropy loss function that measures the difference between the prediction and the ground truth labels.

\noindent\textbf{Local Differential privacy.} In the context of FL, clients collaborate to train a global model under the constraint that each client's data remains local. While this protects the raw data, the gradients shared during training can still leak sensitive information. Traditional DP requires a central trusted party which is often not realistic. To remove that limitation, local differential privacy (LDP) has been proposed. The definition of $(\epsilon, \delta)$-LDP is given as below:

\begin{definition}\label{def1}
A perturbation algorithm $M$ satisfies $(\epsilon, \delta)$-Local Differential Privacy ($(\epsilon, \delta)$-LDP) if, for any pair of adjacent datasets $D$ and $D'$, and for all possible output subsets $S$, the following inequality holds:
\begin{equation}
    Pr[\mathcal{M}(D) \in  S ] \leq e^\epsilon · Pr[\mathcal{M}(D') \in S ] + \delta
\end{equation} 
where $\epsilon$ is the privacy budget of $M$, which quantifies the privacy protection level, and $\delta$ is the probability of the privacy guarantee being violated. A smaller value for $\epsilon$ indicates a smaller gap between two probabilities and thus a stronger privacy.
\end{definition}

\noindent\textbf{Threat Models.} We assume that in the entire federated learning environment, local clients only upload their trained gradients. The central server is intrusted and may initiate a gradient inversion attack while aggregating gradients. This attack maliciously infers the client's private training data by comparing the model broadcast in the previous round with the gradient uploaded by the client in the current round, attempting to restore as much detailed information in $D_i$ as possible. If an eavesdropper in the communication channel intercepts the interaction information between the client and the central server, it also can launch the same attack.

\subsection{FedRE: Protection of Privacy Preference}
The key idea of FedRE is to employ a layer-wise local differential privacy mechanism tailored to local privacy preferences, ensuring precise protection of Privacy-Sensitive Information. More specifically, we first compute the sensitivity of the PSI in the privacy-sensitive region. Then we allocate privacy budgets based on the sensitivity of gradients at different network layers to PSI, reducing the noise from the perturbation of privacy-insensitive information. Moreover, with a novel aggregation mechanism on the server side, FedRE gives priority to the gradients with less sensitive information, minimizing the impact of perturbation on global model performance. The workflow of the proposed framework is shown in Algorithm \ref{alg1} and Fig. \ref{framework} illustrates the FedRE approach. 

\subsubsection{Measure of Sensitive Private Information}\label{sct321}
To quantify the sensitivity of private information of privacy preference, we can re-frame the privacy leakage as an issue of the model gradient's sensitivity to input data. When the model calculates gradients, some parameters may be particularly responsive to changes in the private-sensitive regions. The variability in these parameters could be greater, suggesting that they hold more information from those regions, which could increase the risk of privacy breaches during gradient inversion attacks.  Inspired by this insight, we employ the \textit{Jacobian} matrix of the gradient concerning the input as a tool to gauge the sensitivity of different gradient segments to the input data:
\begin{equation}\label{jaccobian}
    J_l (x)=\frac{\partial g_l(x)}{\partial x}=\frac{\partial}{\partial x} \left [\frac{\partial l(x,y; w)}{\partial w_l}\right ]
\end{equation} 
where $g_l()$ is equivalent to the partial derivative of the loss function $l()$ with respect to the parameters $w$ in the $l$-th layer.
 
Then, we extract the privacy-sensitive regions from the data, and for each pixel within the region, we align the values across different pixel channels with the \textit{frobenius-norm} since Jacobians are compared across layers with different sizes and \textit{frobenius-norm} will consider all dimensions of the data. Assuming the dimensions of the privacy-sensitive region are $(w\times h \times c)$, where $c$ is the number of channels and $w\times h$ represents the region size, we can calculate the aligned sensitivity of privacy-sensitive region $J^{R}_l(x)$:
\begin{equation}\label{aligned}
J^{R}_l (x) = \begin{bmatrix}
\left\| J_l(x)_{\left[a, b, :\right]} \right\|_F & \cdots & \left\| J_l(x)_{\left[a, b+h, :\right]} \right\|_F \\
\vdots & \ddots & \vdots \\
\left\| J_l(x)_{\left[a+w, b, :\right]} \right\|_F & \cdots & \left\| J_l(x)_{\left[a+w, b+h, :\right]} \right\|_F\\
\end{bmatrix}
\end{equation}
Given the client $k$, we compute the average of the aligned sensitivity of the privacy-sensitive region as the PSI score in the $l$-th layer of gradients:
\begin{equation}\label{summed}
    S_l = \frac{1}{w*h}\sum_{i=1}^{w}\sum_{j=1}^{h}J^{R}_l (x)_{\left[i, j\right]}
\end{equation}

\newtcolorbox{mybox1}{
  colback=orange!15!,
  colframe=orange!15!,
  left=0pt,
  top=-2pt,
  bottom=-2pt,
  width=0.80\linewidth
}
\newtcolorbox{mybox2}{
  colback=blue!8!,
  colframe=blue!8!,
  left=0pt,
  top=-2pt,
  bottom=-2pt,
  width=0.80\linewidth
}
\newtcolorbox{mybox3}{
  colback=green!12!,
  colframe=green!12!,
  left=0pt,
  top=-2pt,
  bottom=-2pt,
  width=0.80\linewidth
}

\begin{algorithm}[t]
   \caption{FedRE}
   \label{alg1}
   \SetKwInOut{Input}{Input}
    \Input{$T$: communication round;\ \ $K$: client number;\ \ $\eta$: learning rate;\ \ $\{D_{t}\}_{t=1}^K$: distributed dataset with $K$ clients;\ \ $w$: parameter of the model; $\epsilon_l$: privacy budget for $l$-th layer in model; $\sum \{\epsilon_t\}_{t=1}^{l}$: total privacy budget;\ \ $\mathbb{D}$: the public dataset.} 
Initialize the parameter $w$;\\
    \For{$t=1$ {\bfseries to} $T$}{
         Server randomly selects device subset $S_t$ and send $w$ \\      
        \For{\underline{each selected client} $k \in S_t$ {\rm\bfseries in parallel}}{ 
        \For{each layer $l$ in the local model $w$}{
        \begin{mybox3}
     \textit{\textbf{Measure of Sensitive Private Information}}\\
         Compute the Jacobian matrix $J_l(x)$ of the gradient as the sensitivity with (\ref{jaccobian});\\ 
         Align the sensitivity of
privacy-sensitive region $J^R_l(x)$ from different channels with (\ref{aligned});\\
         Compute the averaged PSI score $S_l$ for each sample with (\ref{summed});\\
      \end{mybox3}
      \vspace{-11pt}
      }
      \vspace{-6pt}
      \begin{mybox1}
         \textit{\textbf{Local Differential Privacy with PSI Score}}\\
         Clip each gradient $g$ to $g^c$ with (\ref{clipping});
         Perturb each gradient $g^c$ with (\ref{mechanism}) and (\ref{delta}).\\
      \end{mybox1}
        Send the model $w^k$ back to the server.\\}
         \underline{{\bfseries At server side}} \\
        \vspace{-4pt}
      \begin{mybox2}
         \textit{\textbf{Parameter Aggregation Mechanism}}\\
         Normalize the weight $\hat{\alpha}$ for the aggregation with the public dataset with (\ref{weight});
         Aggregate the local gradients with the weight to obtain the global model $w$ with (\ref{aggrgate}).
      \end{mybox2}
      \vspace{-11pt}
    }
\end{algorithm}

\subsubsection{Local Differential Privacy with PSI Score}
Acquiring both the gradient $g$ and PSI score $S$, the client starts figuring out the right amount of noise for differential privacy in a layer-wise manner. While a smaller privacy budget represents a stricter protection mechanism, we allocate the privacy budget for each layer according to the PSI score $S_l$, thereby ensuring a balance between the effectiveness of model performance and privacy protection.
\begin{equation}\label{allocate}
    \epsilon_l = \frac{\epsilon \times \frac{1}{S_l}}{\sum_{t=1}^{L} \frac{1}{S_t}}
\end{equation}
Where $\epsilon$ is the total privacy budget set by the client. To implement a differential privacy perturbation mechanism that complies with the privacy budget $\epsilon$ for gradients, we adopt the clipping and noise addition to ensure that the global model update is indistinguishable whether a particular sample is included in the learning process.

\begin{equation}\label{clipping}
    g_l^c = \min\left(1, \frac{C_l}{\|g_l\|}\right)\times g_l
\end{equation}
Where $C_l$ is the clipping threshold of $l$-th layer that controls the maximum contribution of a training sample to global update, $g_l^c$ is the gradient after clipping. Besides, to make it potential attackers to infer specific information of any sample, noise adding is performed to satisfy the randomness requirement of DP. We take the Gaussian mechanism for gradient noise adding to ensure LDP. It adopts $L_2$ norm sensitivity, and adds zero-mean noise with variance ${C_l}^2{\sigma_l}^2\textbf{I}$:
\begin{equation}\label{mechanism}
    \mathcal{M}(g) = g_l^c + \mathcal{N}(0, {C_l}^2{\sigma_l}^2\textbf{I}) 
\end{equation}
Where \textbf{I} is an identity matrix and has the same size with $g_l^c$. $\sigma_l$ is a noise multiplier computed by a privacy accountant and composition mechanism ~\cite{wei2021user} for privacy budget $\epsilon_l$, failure probability $\delta_l$ and communication rounds $T$. 

\begin{equation}\label{delta}
    \sigma_l = \frac{\sqrt{2T\ln(1 / \delta_l)}}{\epsilon_l}  
\end{equation}

\noindent\textbf{Theorem 1 (Simple Composition) } Here we introduce Theorem 1 proposed by ~\cite{kairouz2015composition}. If $\mathcal{M}_{i}$ is an $(\varepsilon_{i}, \delta_{i})$-differentially private (DP) mechanism, then the composition $(\mathcal{M}_{1}, \mathcal{M}_{2}, \ldots, \mathcal{M}_{k})$ satisfies \\ $(\sum_{i=1}^{k} \varepsilon_{i}, \sum_{i=1}^{k} \delta_{i})$-DP.

\vspace{5pt}
\noindent\textbf{Corollary 1 (DP Composition in FedRE)\label{Cor1}} Denote the gradient of the network $w$ with $l$-layers $G=[g_1,g_2,\ldots,g_l]$ , privacy budget for each layer $\epsilon=[\epsilon_1,\epsilon_2,\ldots,\epsilon_l]$, $\sum_{i=1}^{l}\epsilon_i = \epsilon$, and probability of being violated for each layer $\delta=[\delta_1,\delta_2,\ldots,\delta_l]$, $\sum_{i=1}^{l}\delta_i = \delta$. Given any gradient of the $l-th$ layer, the proposed mechanism $M_l$ in \eqref{mechanism} satisfies $(\epsilon_l, \delta_l)$-LDP. Then, the gradient of the $w$ satisfies $(\epsilon, \delta)$-LDP.

\subsubsection{Parameter Aggregation Mechanism}
Considering that the local perturbed gradient will introduce noise to the aggregated global model, which may degrade the model performance, we develop a new Perturbation Distribution Aware Parameter Aggregation Mechanism \textbf{(PDA-PAM)} that can be aware of the distribution of client's parameter perturbation, enabling the server to aggregate the clean gradients (with less perturbation during the model training) from local clients. Assuming that the server has access to the category of local data, then the server employs a public dataset and computes the PSI score of each layer (defined in \ref{sct321}) of each local model $w^k$ with the public dataset $\mathbb{D}$ and then normalizes it into the weight:
\begin{align}\label{weight}
        \hat{\alpha}^{t}_l = \frac{e^{S_l(w^{t}_l;\mathbb{D})}}{\sum_{t=1}^{K}e^{S_l(w^{t}_l;\mathbb{D})}}
\end{align}
which guarantees that $\sum_{t=1}^{K}\hat{\alpha}^{t}=1$. Finally, the server aggregates the local gradients with the PSI score on the public dataset to obtain the global model $w$ for the next communication round:
\begin{equation}\label{aggrgate}
    w_l = \Tilde{w}_l - \frac{\eta}{K}\sum_{t=1}^{K}\frac{g^t_l}{\hat{\alpha}^{t}_l}
\end{equation}
where $\Tilde{w}_l$ denotes the $l$-th layer of the global model in the last communication round. Here the server prefers to aggregate the gradients with less sensitivity thus the global model can gain more effective information.

\subsection{Experimental Results}
\noindent\textbf{Training performance.}
We analyze the utility of different methods on two datasets. As shown in Table \ref{table1}, the complexity and diversity of the DocTamper dataset could pose additional challenges for maintaining high accuracy, especially when the privacy constraint is strict ($\epsilon = 10$). Nonetheless, FedRE consistently performs on par or outperforms other methods across all metrics and datasets, indicating its robustness and adaptability.  
  
Comparing the results across different $\epsilon$ values, we observe that an increase in the privacy budget (i.e., larger $\epsilon$) leads to improved performance for all methods. This is expected, as a larger $\epsilon$ allows for less noise to be injected during the federated learning process, thus facilitating better model convergence.  
  
\textit{The performance gains for FedRE are more notable on the more complex DocTamper dataset when the privacy budget is tight ($\epsilon = 10$).} This underscores the effectiveness of FedRE's PDA-PAM aggregation strategies, which are able to effectively handle the diverse and potentially conflicting perturbations present in the client models' updates. By dynamically adjusting the aggregation based on the specific perturbations, FedRE is able to extract useful information even from heavily distorted updates, demonstrating its resilience in challenging conditions.

\begin{table*}[t]
\caption{Comparison of Training Performance on Different Datasets.}
\label{table1}
\begin{center}
\begin{sc} 
\begin{tabular}{llcccccccc}
\toprule
\multirow{2}{*}{\textbf{$\epsilon$}} & \multirow{2}{*}{\textbf{Methods}} & \multicolumn{4}{c}{\textbf{T-SROIE}} &  \multicolumn{4}{c}{\textbf{DocTamper}} \\
\cmidrule(lr){3-6} \cmidrule(lr){7-10}
& & \textbf{IoU} & \textbf{Precision} & \textbf{Recall} & \textbf{F-Score} & \textbf{IoU} & \textbf{Precision} & \textbf{Recall} & \textbf{F-Score} \\
\midrule
\multirow{2}{*}{$\infty$} 
& Central   & 0.721 \small $\!\pm\!$ \small{0.018} & 0.771 \small $\!\pm\!$ \small{0.008} & 0.917 \small $\!\pm\!$ \small{0.020} & 0.838 \small $\!\pm\!$ \small{0.014} 
& 0.575 \small $\!\pm\!$ \small{0.012} & 0.778 \small $\!\pm\!$ \small{0.006} & 0.688 \small $\!\pm\!$ \small{0.012} & 0.729 \small $\!\pm\!$ \small{0.010} \\
& Localset & 0.408 \small $\!\pm\!$ \small{0.017} & 0.607 \small $\!\pm\!$ \small{0.018} & 0.564 \small $\!\pm\!$ \small{0.016} & 0.585 \small $\!\pm\!$ \small{0.015} 
& 0.345 \small $\!\pm\!$ \small{0.015} & 0.677 \small $\!\pm\!$ \small{0.019} & 0.599 \small $\!\pm\!$ \small{0.014} & 0.635 \small $\!\pm\!$ \small{0.016} \\
\midrule
\multirow{3}{*}{50} 
& LDP-Fed & 0.523 \small $\!\pm\!$ \small{0.019} & 0.654 \small $\!\pm\!$ \small{0.017} & \textbf{0.681\small $\!\pm\!$ \small{0.018}}  & 0.667 \small $\!\pm\!$ \small{0.018} 
& 0.498 \small $\!\pm\!$ \small{0.018} & 0.716 \small $\!\pm\!$ \small{0.018} & \textbf{0.652 \small $\!\pm\!$ \small{0.017}} & 0.682 \small $\!\pm\!$ \small{0.019} \\
& BLUR+LUS & 0.577 \small $\!\pm\!$ \small{0.020} & 0.689 \small $\!\pm\!$ \small{0.019} & 0.639 \small $\!\pm\!$ \small{0.019} & 0.663 \small $\!\pm\!$ \small{0.019} 
& 0.514 \small $\!\pm\!$ \small{0.019} & 0.707 \small $\!\pm\!$ \small{0.020} & 0.626 \small $\!\pm\!$ \small{0.018} & 0.664 \small $\!\pm\!$ \small{0.020} \\
& FedRE & \textbf{0.601} \small $\!\pm\!$ \textbf{\small{0.018}} & \textbf{0.697} \small $\!\pm\!$ \textbf{\small{0.016}} & 0.651 \small $\!\pm\!$ \small{0.017} & \textbf{0.673} \small $\!\pm\!$ \textbf{\small{0.017}} 
& \textbf{0.524} \small $\!\pm\!$ \small{0.017} & \textbf{0.727} \small $\!\pm\!$ \small{0.017} & 0.643 \small $\!\pm\!$ \small{0.016} & \textbf{0.683} \small $\!\pm\!$ \small{0.018} \\
\midrule
\multirow{3}{*}{10} 
& LDP-Fed & 0.430 \small $\!\pm\!$ \small{0.016} & 0.592 \small $\!\pm\!$ \small{0.021} & 0.594 \small $\!\pm\!$ \small{0.015} & 0.593 \small $\!\pm\!$ \small{0.017} 
& 0.424 \small $\!\pm\!$ \small{0.014} & 0.647 \small $\!\pm\!$ \small{0.022} & 0.574 \small $\!\pm\!$ \small{0.013} & 0.608 \small $\!\pm\!$ \small{0.018} \\
& BLUR+LUS & 0.439 \small $\!\pm\!$ \small{0.017} & 0.611 \small $\!\pm\!$ \small{0.023} & 0.606 \small $\!\pm\!$ \small{0.016} & 0.609 \small $\!\pm\!$ \small{0.018} 
& 0.432 \small $\!\pm\!$ \small{0.015} & 0.657 \small $\!\pm\!$ \small{0.024} & 0.583 \small $\!\pm\!$ \small{0.014} & 0.618 \small $\!\pm\!$ \small{0.019} \\
& FedRE & \textbf{0.448} \small $\!\pm\!$ \textbf{\small{0.015}} & \textbf{0.630} \small $\!\pm\!$ \textbf{\small{0.020}} & \textbf{0.618} \small $\!\pm\!$ \textbf{\small{0.014}} & \textbf{0.624} \small $\!\pm\!$ \textbf{\small{0.016}} 
& \textbf{0.457} \small $\!\pm\!$ \small{0.013} & \textbf{0.686} \small $\!\pm\!$ \small{0.021} & \textbf{0.609} \small $\!\pm\!$ \textbf{\small{0.012}} & \textbf{0.645} \small $\!\pm\!$ \textbf{\small{0.017}} \\
\bottomrule
\end{tabular}
\end{sc}
\end{center}
\end{table*}

\begin{table*}[t]
\caption{Comparison of Defense Performance on Different Datasets.}
\label{table2}
\begin{center}
\begin{sc} 
\setlength{\tabcolsep}{2.5pt} 
\begin{tabular}{clcccccccc}
\toprule
\multirow{2}{*}{\textbf{$\epsilon$}} & \multirow{2}{*}{\textbf{Methods}} & \multicolumn{4}{c}{\textbf{T-SROIE}} &  \multicolumn{4}{c}{\textbf{DocTamper}} \\
\cmidrule(lr){3-6} \cmidrule(lr){7-10}
& & \textbf{MSE} $\uparrow$ & \textbf{SSIM} $\downarrow$ & \textbf{PSNR} $\downarrow$ & \textbf{LPIPS} $\uparrow$ & \textbf{MSE} $\uparrow$ & \textbf{SSIM} $\downarrow$ & \textbf{PSNR} $\downarrow$ & \textbf{LPIPS} $\uparrow$ \\
\midrule
\multirow{3}{*}{50} & LDP-Fed & 0.022 {\small $\pm$ 0.003} & 0.937 {\small $\pm$ 0.005} & 48.976 {\small $\pm$ 0.611} & 0.279 {\small $\pm$ 0.014} & 0.027 {\small $\pm$ 0.004} & 0.897 {\small $\pm$ 0.008} & 46.375 {\small $\pm$ 0.693} & 0.288 {\small $\pm$ 0.017} \\
& BLUR+LUS  & 0.020 {\small $\pm$ 0.002} & 0.894 {\small $\pm$ 0.005} & 49.978 {\small $\pm$ 0.520} & 0.269 {\small $\pm$ 0.013} & 0.025 {\small $\pm$ 0.003} & 0.904 {\small $\pm$ 0.007} & 47.877 {\small $\pm$ 0.587} & 0.282 {\small $\pm$ 0.016} \\
& FedRE & \textbf{0.025} {\small $\pm$ \textbf{0.004}} & \textbf{0.808} {\small $\pm$ \textbf{0.008}} & \textbf{47.463} {\small $\pm$ \textbf{0.727}} & \textbf{0.294} {\small $\pm$ \textbf{0.018}} & \textbf{0.030} {\small $\pm$ \textbf{0.004}} & \textbf{0.888} {\small $\pm$ \textbf{0.008}} & \textbf{45.462} {\small $\pm$ \textbf{0.813}} & \textbf{0.303} {\small $\pm$ \textbf{0.019}} \\
\midrule
\multirow{3}{*}{10} & LDP-Fed & 0.327 {\small $\pm$ 0.016} & 0.792 {\small $\pm$ 0.010} & 45.985 {\small $\pm$ 1.211} & 0.391 {\small $\pm$ 0.022} & 0.422 {\small $\pm$ 0.019} & 0.768 {\small $\pm$ 0.011} & 42.986 {\small $\pm$ 1.324} & 0.303 {\small $\pm$ 0.023} \\
& BLUR+LUS  & 0.375 {\small $\pm$ 0.017} & 0.739 {\small $\pm$ 0.009} & 45.987 {\small $\pm$ 1.120} & 0.383 {\small $\pm$ 0.021} & 0.390 {\small $\pm$ 0.011} & 0.776 {\small $\pm$ 0.010} & 43.984 {\small $\pm$ 1.235} & 0.296 {\small $\pm$ 0.022} \\
& FedRE & \textbf{0.383} {\small $\pm$ \textbf{0.014}} & \textbf{0.676} {\small $\pm$ \textbf{0.008}} & \textbf{43.772} {\small $\pm$ \textbf{1.313}} & \textbf{0.412} {\small $\pm$ \textbf{0.024}} & \textbf{0.438} {\small $\pm$ \textbf{0.020}} & \textbf{0.653} {\small $\pm$ \textbf{0.014}} & \textbf{40.367} {\small $\pm$ \textbf{1.421}} & \textbf{0.324} {\small $\pm$ \textbf{0.023}} \\
\bottomrule
\end{tabular}
\end{sc}
\end{center}
\end{table*}

\begin{figure}[t]
\centering
\resizebox{\linewidth}{!}{\includegraphics{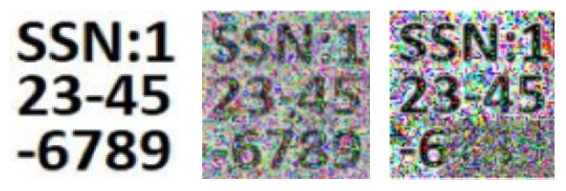}}
\vspace{-0.8cm}
\caption{Images recovered from gradient after gradient leakage attack without FedRE and with FedRE under the same privacy budget. Assuming that the last three characters of an image containing a social security number are PSI, the left image is the original image, the center image is attacked without FedRE, and the right image is the effect of privacy budget reallocation using FedRE. }
\label{recovered demonstratioon}
\vspace{-0.5cm} 
\end{figure}

\vspace{2pt}
\noindent\textbf{Defense performance.}
After we prove that FedRE can provide similar or even superior learning results compared to state-of-the-art DP-based FL mechanisms, we study if the sensitivity computation improves defense ability. In our experimental evaluation, FedRE's sensitivity-driven privacy budget allocation strategy has demonstrated remarkable effectiveness in enhancing the defense capabilities against adversarial attacks in real-world privacy-preserving applications. Specifically, by identifying and prioritizing sensitive personally identifiable information within the data, FedRE is able to allocate more privacy budget to these critical regions. Fig. \ref{recovered demonstratioon} shows an example of a real-life privacy-preserving application of FedRE, e.g., for the same privacy budget, by labeling the last three digits of the social security number as the PSI that need to be protected, FedRE can allocate more privacy budget to the areas that need it, thus successfully blurring the PSI recovered from the attack, and decreasing the likelihood of compromising the privacy information. 
  
Quantitative results presented in Table \ref{table2} consistently show that FedRE outperforms other state-of-the-art DP-based FL mechanisms across various metrics and datasets. This superior performance can be attributed to FedRE's budget allocation strategy, which focuses on protecting sensitive areas more rigorously. Consequently, the similarity between the recovered PSI and the original information is substantially reduced.  
  
\textit{The advantage of FedRE is even more pronounced when the privacy budget ($\epsilon$) is set to a lower value, such as 10. This is because a stricter privacy budget encourages FedRE to allocate more noise to gradients that contain information related to PSI, thereby strengthening the defense. } 
  
The disparity in defense metrics between the T-Sroie and DocTamper datasets highlights FedRE's adaptability to diverse data types. Given that all images in the T-Sroie dataset are grayscale invoices, while Doctamper contains more diverse and colorful images, the defense metrics on the Doctamper dataset exhibit a more pronounced numerical advantage. Nevertheless, FedRE maintains its superior performance, indicating its robustness across different data landscapes.  

\begin{figure}[t]

\centering
\includegraphics[width=\linewidth]{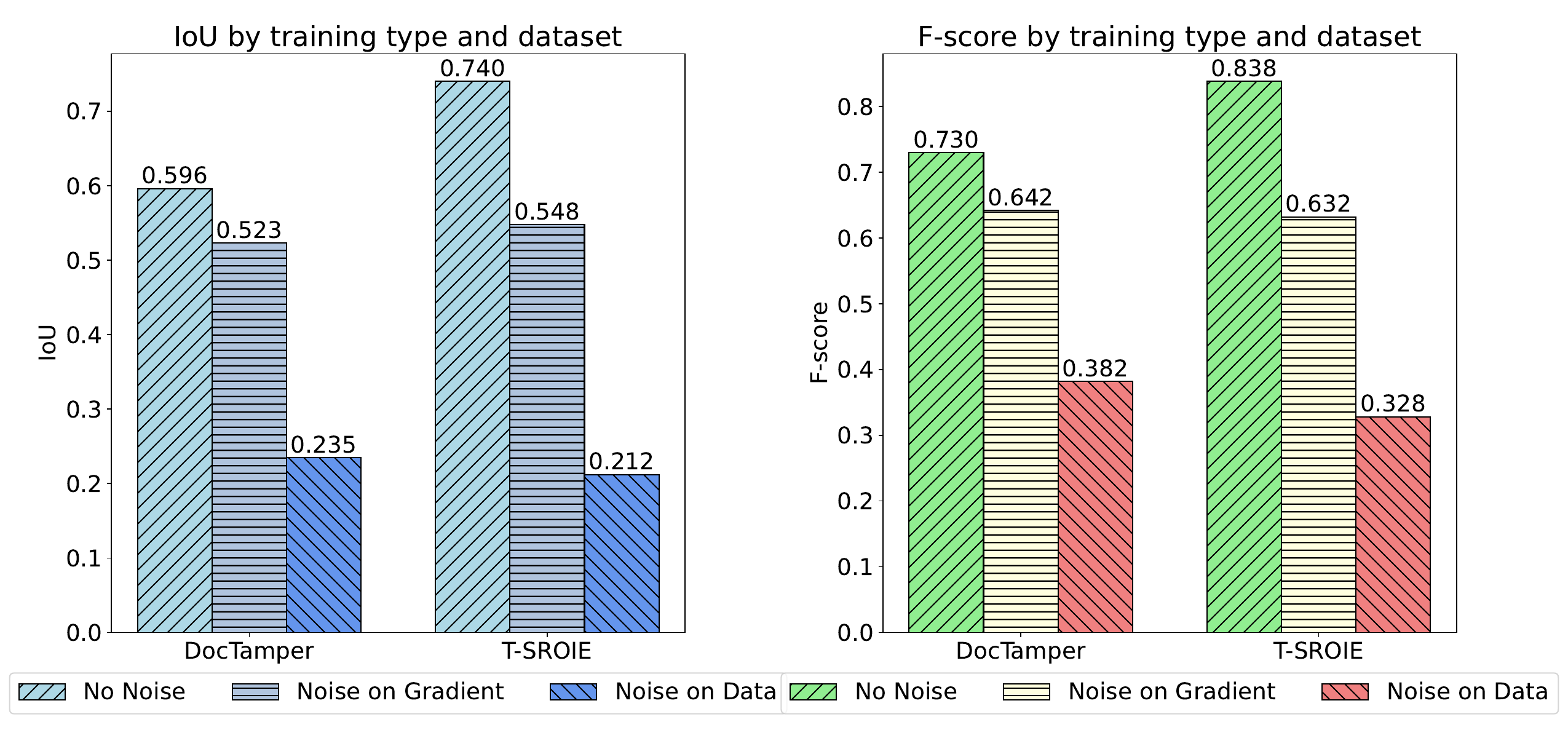}
\vspace{-0.8cm}
\caption{Comparison of the effects of no noise, adding noise to the gradient and adding noise to the raw data on the Doctamper and T-Sroie datasets for training, with the iou metric on the left and the f-score metric on the right.}
\label{comparison on data and gradient}
\vspace{-0.8cm}
\end{figure}

\vspace{2pt}
\noindent\textbf{The Necessity of Gradient Perturbation for PSI Protection.}
To underscore the advantages of perturbing gradients for safeguarding private and sensitive information (PSI), as opposed to directly perturbing raw data at the pixel level, we delve deeper into the utility trade-offs between these two methods. Our initial approach involves directly infusing Gaussian noise into the privacy-sensitive regions of each image, while the alternative approach strategically introduces noise to the gradients during the training process. For a fair comparison, we meticulously adjust the noise intensities to ensure that after 2000 iterations of DLG attack, both methods exhibit comparable defense capabilities, as measured by similarity metrics.   
  
As evident in Fig \ref{comparison on data and gradient}, while gradient perturbation introduces a modest performance decrement, pixel-level perturbation to raw data leads to a drastic deterioration in utility. This disparity stems from the nuanced requirements of tasks such as tamper detection, which heavily rely on intricate noise patterns and accurate color perception in the raw data. Notably, privacy-sensitive regions often coincide with areas that have undergone tampering, thus, applying noise to these overlapping zones disrupts the model's ability to extract meaningful and effective knowledge from them. Consequently, the model's capacity to accurately detect and classify tampering instances is significantly hindered.  
  
The result also shows a dataset-specific trend. The T-SROIE dataset, being relatively smaller in size, appears to be more susceptible to the detrimental effects of noise-augmented training. Specifically, the introduction of noise during training leads to a far more pronounced reduction in IoU and F-score compared to the DocTamper dataset. This observation underscores the importance of tailoring privacy-preserving techniques to the unique characteristics of individual datasets, particularly their size and complexity, to ensure a balanced approach that safeguards privacy without compromising too much on utility.  
  
The gradient perturbation approach offers a more flexible and targeted means of defense. \textit{By perturbing gradients rather than the raw data, we can maintain a higher level of fidelity in the input images, allowing the model to better capture relevant features for downstream tasks.} This targeted intervention not only reduces the overall performance impact but also ensures that the model's ability to detect tampering remains robust, even in the presence of privacy-preserving measures.

\begin{figure}[t]
\centering
\resizebox{\linewidth}{!}{\includegraphics{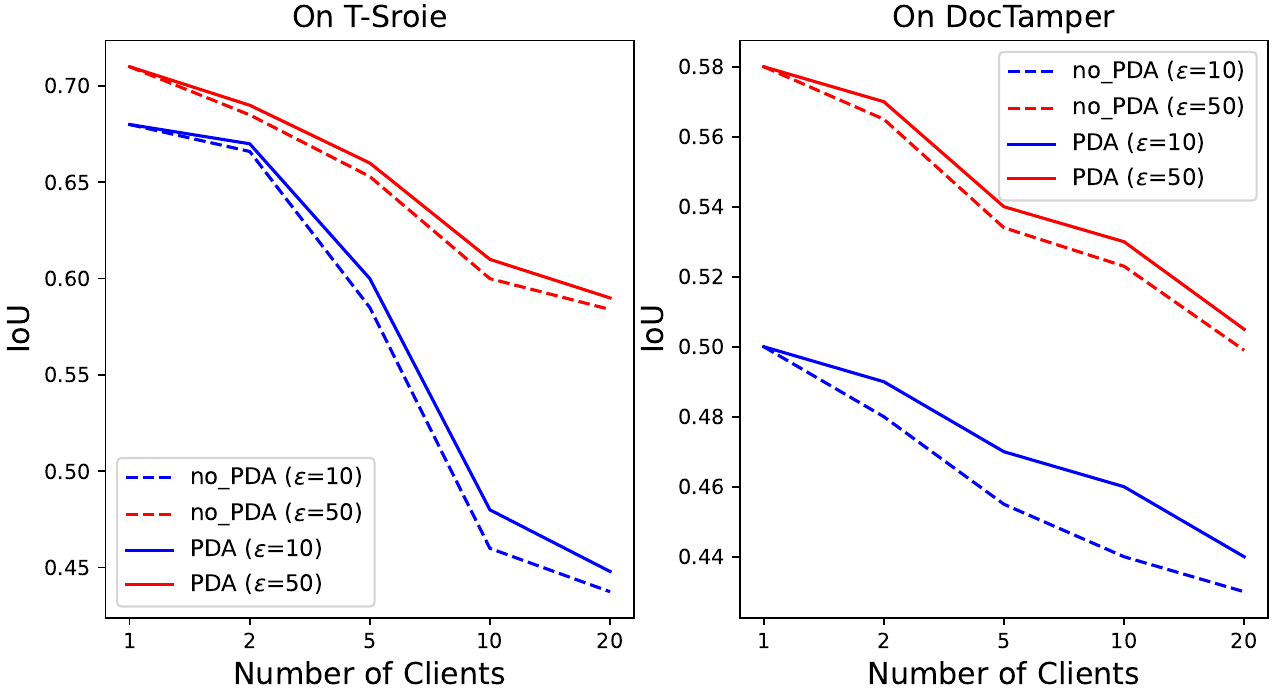}}
\caption{Iou gain of aggregation using PDA-PAM under the different number of clients and privacy overhead settings. }
\label{PDA gain}
\end{figure}

\vspace{2pt}
\noindent\textbf{Parameter Aggregation Mechanism Gain.} Fig. \ref{PDA gain} presents a comparison between IoU results when parameters are aggregated with and without the implementation of PDA-PAM in FedRE. The figure elucidates the impact of varying the privacy budget $\epsilon$ on the performance of the system, particularly under conditions where this budget is constrained.

As the privacy budget $\epsilon$ diminishes, the aggregation of parameters utilizing PDA-PAM from a larger number of clients is observed to compensate more effectively for the noise introduced by the differential privacy constraints. This phenomenon can be attributed to the diversity of Privacy-Sensitive Information (PSI) across different clients. Since the privacy budget $\epsilon_l$ for the same layer may vary among clients, those with a larger $\epsilon_l$ can offer better compensation for clients with a smaller $\epsilon_l$, especially when the collective client count is high. When $\epsilon_l$ is lower, the compensation effect will be more obvious. However, due to the limitation of the fixed size of the data set we use for the experiment, this trend may become less pronounced once the number of clients reaches a certain level. This may be because the amount of information that may be provided by each additional client gradually decreases, which can be mitigated if the data set grows with the number of clients in real-world scenarios.

When the privacy budget is less stringent and the system comprises only a single client, the IoU values surpass those of centralized training without applying differential privacy. This outcome may be accredited to the clipping operation under a lower noise regime, which could potentially enhance gradient regularization. This observation implies that in real life, the addition of differential privacy does not always result in a worse performance.

\vspace{2pt}
\noindent\textbf{Clipping threshold.}
Table \ref{doctamper-Threshold} and \ref{tsroie-Threshold} are experiment results with repect to thresholds $C_l$ in \eqref{clipping} on DocTamper and T-SROIE dataset. All experiments are conducted under the setting of $\epsilon$ = 50, lr = 0.005, and clients number = 10. We can see that a suitable large threshold will not affect the training performance too much, and may even slightly improve performance, acting as a way of regularization. However, too small a threshold will lead to too little information contained in the gradient, resulting in the model being unable to learn during training.

\begin{table}[h]
\caption{DocTamper Dataset Results for Threshold $ C_l $. }
\vspace{-0.2cm}
\label{doctamper-Threshold}
\renewcommand\arraystretch{1.2}
\centering
\scriptsize
\begin{tabular}{ccccc} 
\toprule
$C_l$ & \text{IOU} & \text{PRECISION} & \text{RECALL} & \text{F-SCORE} \\
\midrule
0.20 & 0.535±0.021 & 0.701±0.016 & 0.707±0.017 & 0.702±0.022 \\
0.15 & 0.524±0.017 & 0.727±0.017 & 0.643±0.016 & 0.683±0.018 \\
0.07 & 0.283±0.020 & 0.534±0.016 & 0.344±0.017 & 0.428±0.023 \\
0.05 & 0.260±0.013 & 0.313±0.014 & 0.501±0.015 & 0.384±0.017 \\
0.03 & 0.000 & 0.000 & 0.000 & 0.000 \\
\bottomrule
\vspace{-0.8cm}
\end{tabular}
\end{table}

\begin{table}[h]
\caption{T-SROIE Dataset Results for Threshold $ C_l $. }
\vspace{-0.2cm}
\label{tsroie-Threshold}
\renewcommand\arraystretch{1.2}
\centering
\scriptsize
\begin{tabular}{ccccc} 
\toprule
$C_l$ & \text{IOU} & \text{PRECISION} & \text{RECALL} & \text{F-SCORE} \\
\midrule
0.25 & 0.593±0.020 & 0.687±0.020 & 0.649±0.018 & 0.671±0.021 \\
0.20 & 0.601±0.018 & 0.697±0.016 & 0.651±0.017 & 0.673±0.017 \\
0.15 & 0.322±0.017 & 0.378±0.018 & 0.623±0.015 & 0.465±0.015 \\
0.14 & 0.311±0.018 & 0.336±0.018 & 0.601±0.016 & 0.435±0.014 \\
0.13 & 0.000 & 0.000 & 0.000 & 0.000 \\
\bottomrule
\end{tabular}
\end{table}

\vspace{2pt}
\noindent\textbf{Computational Overhead Analysis.} The process of calculating the PSI score involves three processes, corresponding to \eqref{jaccobian}, \eqref{aligned}, and \eqref{summed} in the paper.

Equation \eqref{jaccobian} corresponds to the process of evaluating the Jacobian matrix, which involves two sub steps; the first step is to obtain the gradient by calculate derivative of the loss function with respect to the model weights, and the second step is to obtain the Jacobian matrix by taking the derivative of gradient with respect to the input data of the model. The time complexity of the first step is $O(F)$, $ F$ represents the total number of floating-point computations performed by the model, and the space complexity is $O(N)$, $ N$ represents the number of parameters of the model. This step is the same as the original gradient computation process in model training task, and thus can utilize the intermediate results generated during training without incurring any additional time and space overheads. The time complexity of the second step is $O(FD)$, where $D$ is the number of features in the input data and the space complexity is $O(DN)$. Equation \eqref{aligned} corresponds to aligning the sensitivity of the privacy-sensitive region, assuming that the total number of pixel points in the privacy-sensitive region is $p$, the time complexity is $O(p)$ and the space complexity is $O(p)$. Equation \eqref{summed} corresponds to the calculation of the average PSI score within the privacy-sensitive region, with a time complexity of $O(p)$ and a space complexity of $O(p)$.

\textit{$p$ is generally much smaller than $FD$ and $DN$, so in summary the time complexity of the algorithm is $O(FD)$ and the space complexity is $O(DN)$.}

While the calculation of PSI scores has some overhead, \textit{in practice it is not necessary to calculate PSI scores for every training data, but only when dealing with data with different content formats.} Financial data generally have several fixed formats, such as contract, invoice, normal page, receipt, etc.. There are large differences in the data formats between different image layouts, and therefore, the PSI scores vary widely. Data in the same format have similar PSI scores due to the same image layout, similar privacy protection preferences, and tampering locations. In practice, the average of the PSI scores calculated by sampling 10 data in the same format is used instead of the PSI of all the training data.

\section{Conclusion and Future Work}
In this paper, we propose a federated mechanism called FedRE that can simultaneously achieve robustness and effectiveness benefits with LDP protection. It considers different privacy preferences on privacy-sensitive information of clients, perturbs the parameters adaptively, and aggregates parameters based on the distribution of perturbed information. It not only achieves better privacy protection, decreasing the similarity between the reconstructed images and raw images in sensitive regions, but also reduces the noise when aggregating and improves the performance of the model.

While FedRE presents a robust framework for federated learning with privacy preservation, there are areas that merit further exploration and improvement. In future research, PSI computational efficiency can be further enhanced by employing optimization methods like utilizing the sparsity of the Jacobian matrix and leveraging approximate computation methods. Additionally, the exploration of optimizing the sampling computation of PSI scores to approximate the overall distribution effectively, especially in scenarios characterized by high data heterogeneity and imbalance, is also a noteworthy area of investigation.

\begin{acks}
This work is supported by the National Key Research and Development Program of China under grant 2024YFC3307900; the National Natural Science Foundation of China under grants 62376103, 62302184, 62436003 and 62206102; Major Science and Technology Project of Hubei Province under grant 2024BAA008; Hubei Science and Technology Talent Service Project under grant 2024DJC078; and Ant Group through CCF-Ant Research Fund. The computation is completed in the HPC Platform of Huazhong University of Science and Technology.  
\end{acks}

\bibliographystyle{ACM-Reference-Format}
\bibliography{sample-base}


\begin{thebibliography}{48}


\ifx \showCODEN    \undefined \def \showCODEN     #1{\unskip}     \fi
\ifx \showISBNx    \undefined \def \showISBNx     #1{\unskip}     \fi
\ifx \showISBNxiii \undefined \def \showISBNxiii  #1{\unskip}     \fi
\ifx \showISSN     \undefined \def \showISSN      #1{\unskip}     \fi
\ifx \showLCCN     \undefined \def \showLCCN      #1{\unskip}     \fi
\ifx \shownote     \undefined \def \shownote      #1{#1}          \fi
\ifx \showarticletitle \undefined \def \showarticletitle #1{#1}   \fi
\ifx \showURL      \undefined \def \showURL       {\relax}        \fi
\providecommand\bibfield[2]{#2}
\providecommand\bibinfo[2]{#2}
\providecommand\natexlab[1]{#1}
\providecommand\showeprint[2][]{arXiv:#2}

\bibitem[Antunes et~al\mbox{.}(2022)]%
        {Antunes22}
\bibfield{author}{\bibinfo{person}{Rodolfo~Stoffel Antunes}, \bibinfo{person}{Cristiano Andr{\'e}~da Costa}, \bibinfo{person}{Arne K{\"u}derle}, \bibinfo{person}{Imrana~Abdullahi Yari}, {and} \bibinfo{person}{Bj{\"o}rn Eskofier}.} \bibinfo{year}{2022}\natexlab{}.
\newblock \showarticletitle{Federated learning for healthcare: Systematic review and architecture proposal}.
\newblock \bibinfo{journal}{\emph{ACM Transactions on Intelligent Systems and Technology (TIST)}} \bibinfo{volume}{13}, \bibinfo{number}{4} (\bibinfo{year}{2022}), \bibinfo{pages}{1--23}.
\newblock


\bibitem[Arachchige et~al\mbox{.}(2019)]%
        {arachchige2019local}
\bibfield{author}{\bibinfo{person}{Pathum Chamikara~Mahawaga Arachchige}, \bibinfo{person}{Peter Bertok}, \bibinfo{person}{Ibrahim Khalil}, \bibinfo{person}{Dongxi Liu}, \bibinfo{person}{Seyit Camtepe}, {and} \bibinfo{person}{Mohammed Atiquzzaman}.} \bibinfo{year}{2019}\natexlab{}.
\newblock \showarticletitle{Local differential privacy for deep learning}.
\newblock \bibinfo{journal}{\emph{IEEE Internet of Things Journal}} \bibinfo{volume}{7}, \bibinfo{number}{7} (\bibinfo{year}{2019}), \bibinfo{pages}{5827--5842}.
\newblock


\bibitem[Bonawitz et~al\mbox{.}(2017)]%
        {Bonawitz17}
\bibfield{author}{\bibinfo{person}{Keith Bonawitz}, \bibinfo{person}{Vladimir Ivanov}, \bibinfo{person}{Ben Kreuter}, \bibinfo{person}{Antonio Marcedone}, \bibinfo{person}{H~Brendan McMahan}, \bibinfo{person}{Sarvar Patel}, \bibinfo{person}{Daniel Ramage}, \bibinfo{person}{Aaron Segal}, {and} \bibinfo{person}{Karn Seth}.} \bibinfo{year}{2017}\natexlab{}.
\newblock \showarticletitle{Practical secure aggregation for privacy-preserving machine learning}. In \bibinfo{booktitle}{\emph{proceedings of the 2017 ACM SIGSAC Conference on Computer and Communications Security}}. \bibinfo{pages}{1175--1191}.
\newblock


\bibitem[Byrd and Polychroniadou(2020)]%
        {byrd2020}
\bibfield{author}{\bibinfo{person}{David Byrd} {and} \bibinfo{person}{Antigoni Polychroniadou}.} \bibinfo{year}{2020}\natexlab{}.
\newblock \showarticletitle{Differentially private secure multi-party computation for federated learning in financial applications}. In \bibinfo{booktitle}{\emph{Proceedings of the First ACM International Conference on AI in Finance}}. \bibinfo{pages}{1--9}.
\newblock


\bibitem[Cheng et~al\mbox{.}(2022)]%
        {cheng2022}
\bibfield{author}{\bibinfo{person}{Anda Cheng}, \bibinfo{person}{Peisong Wang}, \bibinfo{person}{Xi~Sheryl Zhang}, {and} \bibinfo{person}{Jian Cheng}.} \bibinfo{year}{2022}\natexlab{}.
\newblock \showarticletitle{Differentially private federated learning with local regularization and sparsification}. In \bibinfo{booktitle}{\emph{Proceedings of the IEEE/CVF Conference on Computer Vision and Pattern Recognition}}. \bibinfo{pages}{10122--10131}.
\newblock


\bibitem[Cormode et~al\mbox{.}(2018)]%
        {cormode2018privacy}
\bibfield{author}{\bibinfo{person}{Graham Cormode}, \bibinfo{person}{Somesh Jha}, \bibinfo{person}{Tejas Kulkarni}, \bibinfo{person}{Ninghui Li}, \bibinfo{person}{Divesh Srivastava}, {and} \bibinfo{person}{Tianhao Wang}.} \bibinfo{year}{2018}\natexlab{}.
\newblock \showarticletitle{Privacy at scale: Local differential privacy in practice}. In \bibinfo{booktitle}{\emph{Proceedings of the 2018 International Conference on Management of Data}}. \bibinfo{pages}{1655--1658}.
\newblock


\bibitem[Deng et~al\mbox{.}(2023)]%
        {deng2023federated}
\bibfield{author}{\bibinfo{person}{Tianchi Deng}, \bibinfo{person}{Yingguang Li}, \bibinfo{person}{Xu Liu}, {and} \bibinfo{person}{Lihui Wang}.} \bibinfo{year}{2023}\natexlab{}.
\newblock \showarticletitle{Federated learning-based collaborative manufacturing for complex parts}.
\newblock \bibinfo{journal}{\emph{Journal of Intelligent Manufacturing}} \bibinfo{volume}{34}, \bibinfo{number}{7} (\bibinfo{year}{2023}), \bibinfo{pages}{3025--3038}.
\newblock


\bibitem[Geiping et~al\mbox{.}(2020)]%
        {Geiping20}
\bibfield{author}{\bibinfo{person}{Jonas Geiping}, \bibinfo{person}{Hartmut Bauermeister}, \bibinfo{person}{Hannah Dr{\"o}ge}, {and} \bibinfo{person}{Michael Moeller}.} \bibinfo{year}{2020}\natexlab{}.
\newblock \showarticletitle{Inverting gradients-how easy is it to break privacy in federated learning?}
\newblock \bibinfo{journal}{\emph{Advances in Neural Information Processing Systems}}  \bibinfo{volume}{33} (\bibinfo{year}{2020}), \bibinfo{pages}{16937--16947}.
\newblock


\bibitem[Gu et~al\mbox{.}(2023)]%
        {gu2023dynamic}
\bibfield{author}{\bibinfo{person}{Zishan Gu}, \bibinfo{person}{Ke Zhang}, \bibinfo{person}{Guangji Bai}, \bibinfo{person}{Liang Chen}, \bibinfo{person}{Liang Zhao}, {and} \bibinfo{person}{Carl Yang}.} \bibinfo{year}{2023}\natexlab{}.
\newblock \showarticletitle{Dynamic activation of clients and parameters for federated learning over heterogeneous graphs}. In \bibinfo{booktitle}{\emph{2023 IEEE 39th International Conference on Data Engineering (ICDE)}}. IEEE, \bibinfo{pages}{1597--1610}.
\newblock


\bibitem[Huang et~al\mbox{.}(2021)]%
        {Huang21}
\bibfield{author}{\bibinfo{person}{Yangsibo Huang}, \bibinfo{person}{Samyak Gupta}, \bibinfo{person}{Zhao Song}, \bibinfo{person}{Kai Li}, {and} \bibinfo{person}{Sanjeev Arora}.} \bibinfo{year}{2021}\natexlab{}.
\newblock \showarticletitle{Evaluating gradient inversion attacks and defenses in federated learning}.
\newblock \bibinfo{journal}{\emph{Advances in Neural Information Processing Systems}}  \bibinfo{volume}{34} (\bibinfo{year}{2021}), \bibinfo{pages}{7232--7241}.
\newblock


\bibitem[Jeon et~al\mbox{.}(2021)]%
        {Jeon21}
\bibfield{author}{\bibinfo{person}{Jinwoo Jeon}, \bibinfo{person}{Kangwook Lee}, \bibinfo{person}{Sewoong Oh}, \bibinfo{person}{Jungseul Ok}, {et~al\mbox{.}}} \bibinfo{year}{2021}\natexlab{}.
\newblock \showarticletitle{Gradient inversion with generative image prior}.
\newblock \bibinfo{journal}{\emph{Advances in neural information processing systems}}  \bibinfo{volume}{34} (\bibinfo{year}{2021}), \bibinfo{pages}{29898--29908}.
\newblock


\bibitem[Kairouz et~al\mbox{.}(2021)]%
        {kairouz2021advances}
\bibfield{author}{\bibinfo{person}{Peter Kairouz}, \bibinfo{person}{H~Brendan McMahan}, \bibinfo{person}{Brendan Avent}, \bibinfo{person}{Aur{\'e}lien Bellet}, \bibinfo{person}{Mehdi Bennis}, \bibinfo{person}{Arjun~Nitin Bhagoji}, \bibinfo{person}{Kallista Bonawitz}, \bibinfo{person}{Zachary Charles}, \bibinfo{person}{Graham Cormode}, \bibinfo{person}{Rachel Cummings}, {et~al\mbox{.}}} \bibinfo{year}{2021}\natexlab{}.
\newblock \showarticletitle{Advances and open problems in federated learning}.
\newblock \bibinfo{journal}{\emph{Foundations and Trends{\textregistered} in Machine Learning}} \bibinfo{volume}{14}, \bibinfo{number}{1--2} (\bibinfo{year}{2021}), \bibinfo{pages}{1--210}.
\newblock


\bibitem[Kairouz et~al\mbox{.}(2015)]%
        {kairouz2015composition}
\bibfield{author}{\bibinfo{person}{Peter Kairouz}, \bibinfo{person}{Sewoong Oh}, {and} \bibinfo{person}{Pramod Viswanath}.} \bibinfo{year}{2015}\natexlab{}.
\newblock \showarticletitle{The composition theorem for differential privacy}. In \bibinfo{booktitle}{\emph{International conference on machine learning}}. PMLR, \bibinfo{pages}{1376--1385}.
\newblock


\bibitem[Khan et~al\mbox{.}(2020)]%
        {khan2020resource}
\bibfield{author}{\bibinfo{person}{Latif~U Khan}, \bibinfo{person}{Madyan Alsenwi}, \bibinfo{person}{Ibrar Yaqoob}, \bibinfo{person}{Muhammad Imran}, \bibinfo{person}{Zhu Han}, {and} \bibinfo{person}{Choong~Seon Hong}.} \bibinfo{year}{2020}\natexlab{}.
\newblock \showarticletitle{Resource optimized federated learning-enabled cognitive internet of things for smart industries}.
\newblock \bibinfo{journal}{\emph{IEEE Access}}  \bibinfo{volume}{8} (\bibinfo{year}{2020}), \bibinfo{pages}{168854--168864}.
\newblock


\bibitem[Li et~al\mbox{.}(2022)]%
        {li2022ressfl}
\bibfield{author}{\bibinfo{person}{Jingtao Li}, \bibinfo{person}{Adnan~Siraj Rakin}, \bibinfo{person}{Xing Chen}, \bibinfo{person}{Zhezhi He}, \bibinfo{person}{Deliang Fan}, {and} \bibinfo{person}{Chaitali Chakrabarti}.} \bibinfo{year}{2022}\natexlab{}.
\newblock \showarticletitle{Ressfl: A resistance transfer framework for defending model inversion attack in split federated learning}. In \bibinfo{booktitle}{\emph{Proceedings of the IEEE/CVF Conference on Computer Vision and Pattern Recognition}}. \bibinfo{pages}{10194--10202}.
\newblock


\bibitem[Li et~al\mbox{.}(2020)]%
        {li2020federated}
\bibfield{author}{\bibinfo{person}{Tian Li}, \bibinfo{person}{Anit~Kumar Sahu}, \bibinfo{person}{Ameet Talwalkar}, {and} \bibinfo{person}{Virginia Smith}.} \bibinfo{year}{2020}\natexlab{}.
\newblock \showarticletitle{Federated learning: Challenges, methods, and future directions}.
\newblock \bibinfo{journal}{\emph{IEEE signal processing magazine}} \bibinfo{volume}{37}, \bibinfo{number}{3} (\bibinfo{year}{2020}), \bibinfo{pages}{50--60}.
\newblock


\bibitem[Li et~al\mbox{.}(2024)]%
        {li2024towards}
\bibfield{author}{\bibinfo{person}{Yichen Li}, \bibinfo{person}{Qunwei Li}, \bibinfo{person}{Haozhao Wang}, \bibinfo{person}{Ruixuan Li}, \bibinfo{person}{Wenliang Zhong}, {and} \bibinfo{person}{Guannan Zhang}.} \bibinfo{year}{2024}\natexlab{}.
\newblock \showarticletitle{Towards Efficient Replay in Federated Incremental Learning}. In \bibinfo{booktitle}{\emph{Proceedings of the IEEE/CVF Conference on Computer Vision and Pattern Recognition}}. \bibinfo{pages}{12820--12829}.
\newblock


\bibitem[Li et~al\mbox{.}(2023)]%
        {Li23}
\bibfield{author}{\bibinfo{person}{Zhaohua Li}, \bibinfo{person}{Le Wang}, \bibinfo{person}{Guangyao Chen}, \bibinfo{person}{Muhammad Shafq}, {et~al\mbox{.}}} \bibinfo{year}{2023}\natexlab{}.
\newblock \showarticletitle{A survey of image gradient inversion against federated learning}.
\newblock \bibinfo{journal}{\emph{Authorea Preprints}} (\bibinfo{year}{2023}).
\newblock


\bibitem[Liu et~al\mbox{.}(2022)]%
        {Liu22}
\bibfield{author}{\bibinfo{person}{Ji Liu}, \bibinfo{person}{Jizhou Huang}, \bibinfo{person}{Yang Zhou}, \bibinfo{person}{Xuhong Li}, \bibinfo{person}{Shilei Ji}, \bibinfo{person}{Haoyi Xiong}, {and} \bibinfo{person}{Dejing Dou}.} \bibinfo{year}{2022}\natexlab{}.
\newblock \showarticletitle{From distributed machine learning to federated learning: A survey}.
\newblock \bibinfo{journal}{\emph{Knowledge and Information Systems}} \bibinfo{volume}{64}, \bibinfo{number}{4} (\bibinfo{year}{2022}), \bibinfo{pages}{885--917}.
\newblock


\bibitem[Liu et~al\mbox{.}(2023)]%
        {liu2023}
\bibfield{author}{\bibinfo{person}{Ruixuan Liu}, \bibinfo{person}{Yang Cao}, \bibinfo{person}{Yanlin Wang}, \bibinfo{person}{Lingjuan Lyu}, \bibinfo{person}{Yun Chen}, {and} \bibinfo{person}{Hong Chen}.} \bibinfo{year}{2023}\natexlab{}.
\newblock \showarticletitle{PrivateRec: Differentially Private Model Training and Online Serving for Federated News Recommendation}. In \bibinfo{booktitle}{\emph{Proceedings of the 29th ACM SIGKDD Conference on Knowledge Discovery and Data Mining}}. \bibinfo{pages}{4539--4548}.
\newblock


\bibitem[Long et~al\mbox{.}(2020)]%
        {Long2020}
\bibfield{author}{\bibinfo{person}{Guodong Long}, \bibinfo{person}{Yue Tan}, \bibinfo{person}{Jing Jiang}, {and} \bibinfo{person}{Chengqi Zhang}.} \bibinfo{year}{2020}\natexlab{}.
\newblock \showarticletitle{Federated learning for open banking}.
\newblock In \bibinfo{booktitle}{\emph{Federated Learning: Privacy and Incentive}}. \bibinfo{publisher}{Springer}, \bibinfo{pages}{240--254}.
\newblock


\bibitem[Luopan et~al\mbox{.}(2023)]%
        {luopan2023fedknow}
\bibfield{author}{\bibinfo{person}{Yaxin Luopan}, \bibinfo{person}{Rui Han}, \bibinfo{person}{Qinglong Zhang}, \bibinfo{person}{Chi~Harold Liu}, \bibinfo{person}{Guoren Wang}, {and} \bibinfo{person}{Lydia~Y Chen}.} \bibinfo{year}{2023}\natexlab{}.
\newblock \showarticletitle{Fedknow: Federated continual learning with signature task knowledge integration at edge}. In \bibinfo{booktitle}{\emph{2023 IEEE 39th International Conference on Data Engineering (ICDE)}}. IEEE, \bibinfo{pages}{341--354}.
\newblock


\bibitem[McMahan et~al\mbox{.}(2017)]%
        {Mcmahan17}
\bibfield{author}{\bibinfo{person}{Brendan McMahan}, \bibinfo{person}{Eider Moore}, \bibinfo{person}{Daniel Ramage}, \bibinfo{person}{Seth Hampson}, {and} \bibinfo{person}{Blaise~Aguera y Arcas}.} \bibinfo{year}{2017}\natexlab{}.
\newblock \showarticletitle{Communication-efficient learning of deep networks from decentralized data}. In \bibinfo{booktitle}{\emph{Artificial intelligence and statistics}}. PMLR, \bibinfo{pages}{1273--1282}.
\newblock


\bibitem[Pan et~al\mbox{.}(2023)]%
        {pan2023lumos}
\bibfield{author}{\bibinfo{person}{Qiying Pan}, \bibinfo{person}{Yifei Zhu}, {and} \bibinfo{person}{Lingyang Chu}.} \bibinfo{year}{2023}\natexlab{}.
\newblock \showarticletitle{Lumos: Heterogeneity-aware federated graph learning over decentralized devices}. In \bibinfo{booktitle}{\emph{2023 IEEE 39th International Conference on Data Engineering (ICDE)}}. IEEE, \bibinfo{pages}{1914--1926}.
\newblock


\bibitem[Qin et~al\mbox{.}(2023a)]%
        {qin2023fedapen}
\bibfield{author}{\bibinfo{person}{Zhen Qin}, \bibinfo{person}{Shuiguang Deng}, \bibinfo{person}{Mingyu Zhao}, {and} \bibinfo{person}{Xueqiang Yan}.} \bibinfo{year}{2023}\natexlab{a}.
\newblock \showarticletitle{FedAPEN: Personalized Cross-silo Federated Learning with Adaptability to Statistical Heterogeneity}. In \bibinfo{booktitle}{\emph{Proceedings of the 29th ACM SIGKDD Conference on Knowledge Discovery and Data Mining}}. \bibinfo{pages}{1954--1964}.
\newblock


\bibitem[Qin et~al\mbox{.}(2023b)]%
        {qin2023revisiting}
\bibfield{author}{\bibinfo{person}{Zeyu Qin}, \bibinfo{person}{Liuyi Yao}, \bibinfo{person}{Daoyuan Chen}, \bibinfo{person}{Yaliang Li}, \bibinfo{person}{Bolin Ding}, {and} \bibinfo{person}{Minhao Cheng}.} \bibinfo{year}{2023}\natexlab{b}.
\newblock \showarticletitle{Revisiting Personalized Federated Learning: Robustness Against Backdoor Attacks}.
\newblock \bibinfo{journal}{\emph{arXiv preprint arXiv:2302.01677}} (\bibinfo{year}{2023}).
\newblock


\bibitem[Rahman and Purushotham(2023)]%
        {Rahman2023}
\bibfield{author}{\bibinfo{person}{Md~Mahmudur Rahman} {and} \bibinfo{person}{Sanjay Purushotham}.} \bibinfo{year}{2023}\natexlab{}.
\newblock \showarticletitle{FedPseudo: Privacy-Preserving Pseudo Value-Based Deep Learning Models for Federated Survival Analysis}. In \bibinfo{booktitle}{\emph{Proceedings of the 29th ACM SIGKDD Conference on Knowledge Discovery and Data Mining}}. \bibinfo{pages}{1999--2009}.
\newblock


\bibitem[Rieke et~al\mbox{.}(2020)]%
        {rieke2020future}
\bibfield{author}{\bibinfo{person}{Nicola Rieke}, \bibinfo{person}{Jonny Hancox}, \bibinfo{person}{Wenqi Li}, \bibinfo{person}{Fausto Milletari}, \bibinfo{person}{Holger~R Roth}, \bibinfo{person}{Shadi Albarqouni}, \bibinfo{person}{Spyridon Bakas}, \bibinfo{person}{Mathieu~N Galtier}, \bibinfo{person}{Bennett~A Landman}, \bibinfo{person}{Klaus Maier-Hein}, {et~al\mbox{.}}} \bibinfo{year}{2020}\natexlab{}.
\newblock \showarticletitle{The future of digital health with federated learning}.
\newblock \bibinfo{journal}{\emph{NPJ digital medicine}} \bibinfo{volume}{3}, \bibinfo{number}{1} (\bibinfo{year}{2020}), \bibinfo{pages}{119}.
\newblock


\bibitem[Sun et~al\mbox{.}(2023)]%
        {sun2023shapleyfl}
\bibfield{author}{\bibinfo{person}{Qiheng Sun}, \bibinfo{person}{Xiang Li}, \bibinfo{person}{Jiayao Zhang}, \bibinfo{person}{Li Xiong}, \bibinfo{person}{Weiran Liu}, \bibinfo{person}{Jinfei Liu}, \bibinfo{person}{Zhan Qin}, {and} \bibinfo{person}{Kui Ren}.} \bibinfo{year}{2023}\natexlab{}.
\newblock \showarticletitle{Shapleyfl: Robust federated learning based on shapley value}. In \bibinfo{booktitle}{\emph{Proceedings of the 29th ACM SIGKDD Conference on Knowledge Discovery and Data Mining}}. \bibinfo{pages}{2096--2108}.
\newblock


\bibitem[Truex et~al\mbox{.}(2020)]%
        {Truex20}
\bibfield{author}{\bibinfo{person}{Stacey Truex}, \bibinfo{person}{Ling Liu}, \bibinfo{person}{Ka-Ho Chow}, \bibinfo{person}{Mehmet~Emre Gursoy}, {and} \bibinfo{person}{Wenqi Wei}.} \bibinfo{year}{2020}\natexlab{}.
\newblock \showarticletitle{LDP-Fed: Federated learning with local differential privacy}. In \bibinfo{booktitle}{\emph{Proceedings of the Third ACM International Workshop on Edge Systems, Analytics and Networking}}. \bibinfo{pages}{61--66}.
\newblock


\bibitem[Wang et~al\mbox{.}(2023)]%
        {wang2023dafkd}
\bibfield{author}{\bibinfo{person}{Haozhao Wang}, \bibinfo{person}{Yichen Li}, \bibinfo{person}{Wenchao Xu}, \bibinfo{person}{Ruixuan Li}, \bibinfo{person}{Yufeng Zhan}, {and} \bibinfo{person}{Zhigang Zeng}.} \bibinfo{year}{2023}\natexlab{}.
\newblock \showarticletitle{DaFKD: Domain-aware Federated Knowledge Distillation}. In \bibinfo{booktitle}{\emph{Proceedings of the IEEE/CVF Conference on Computer Vision and Pattern Recognition}}. \bibinfo{pages}{20412--20421}.
\newblock


\bibitem[Wang et~al\mbox{.}(2022)]%
        {Wang22}
\bibfield{author}{\bibinfo{person}{Junxiao Wang}, \bibinfo{person}{Song Guo}, \bibinfo{person}{Xin Xie}, {and} \bibinfo{person}{Heng Qi}.} \bibinfo{year}{2022}\natexlab{}.
\newblock \showarticletitle{Protect privacy from gradient leakage attack in federated learning}. In \bibinfo{booktitle}{\emph{IEEE INFOCOM 2022-IEEE Conference on Computer Communications}}. IEEE, \bibinfo{pages}{580--589}.
\newblock


\bibitem[Wei et~al\mbox{.}(2021)]%
        {wei2021user}
\bibfield{author}{\bibinfo{person}{Kang Wei}, \bibinfo{person}{Jun Li}, \bibinfo{person}{Ming Ding}, \bibinfo{person}{Chuan Ma}, \bibinfo{person}{Hang Su}, \bibinfo{person}{Bo Zhang}, {and} \bibinfo{person}{H~Vincent Poor}.} \bibinfo{year}{2021}\natexlab{}.
\newblock \showarticletitle{User-level privacy-preserving federated learning: Analysis and performance optimization}.
\newblock \bibinfo{journal}{\emph{IEEE Transactions on Mobile Computing}} \bibinfo{volume}{21}, \bibinfo{number}{9} (\bibinfo{year}{2021}), \bibinfo{pages}{3388--3401}.
\newblock


\bibitem[Wei et~al\mbox{.}(2020a)]%
        {Wei20}
\bibfield{author}{\bibinfo{person}{Kang Wei}, \bibinfo{person}{Jun Li}, \bibinfo{person}{Ming Ding}, \bibinfo{person}{Chuan Ma}, \bibinfo{person}{Howard~H Yang}, \bibinfo{person}{Farhad Farokhi}, \bibinfo{person}{Shi Jin}, \bibinfo{person}{Tony~QS Quek}, {and} \bibinfo{person}{H~Vincent Poor}.} \bibinfo{year}{2020}\natexlab{a}.
\newblock \showarticletitle{Federated learning with differential privacy: Algorithms and performance analysis}.
\newblock \bibinfo{journal}{\emph{IEEE Transactions on Information Forensics and Security}}  \bibinfo{volume}{15} (\bibinfo{year}{2020}), \bibinfo{pages}{3454--3469}.
\newblock


\bibitem[Wei et~al\mbox{.}(2020b)]%
        {wei2020framework}
\bibfield{author}{\bibinfo{person}{Wenqi Wei}, \bibinfo{person}{Ling Liu}, \bibinfo{person}{Margaret Loper}, \bibinfo{person}{Ka-Ho Chow}, \bibinfo{person}{Mehmet~Emre Gursoy}, \bibinfo{person}{Stacey Truex}, {and} \bibinfo{person}{Yanzhao Wu}.} \bibinfo{year}{2020}\natexlab{b}.
\newblock \showarticletitle{A framework for evaluating gradient leakage attacks in federated learning}.
\newblock \bibinfo{journal}{\emph{arXiv preprint arXiv:2004.10397}} (\bibinfo{year}{2020}).
\newblock


\bibitem[Wu et~al\mbox{.}(2023)]%
        {wu2023serverless}
\bibfield{author}{\bibinfo{person}{Xidong Wu}, \bibinfo{person}{Zhengmian Hu}, \bibinfo{person}{Jian Pei}, {and} \bibinfo{person}{Heng Huang}.} \bibinfo{year}{2023}\natexlab{}.
\newblock \showarticletitle{Serverless federated auprc optimization for multi-party collaborative imbalanced data mining}. In \bibinfo{booktitle}{\emph{Proceedings of the 29th ACM SIGKDD conference on knowledge discovery and data mining}}. \bibinfo{pages}{2648--2659}.
\newblock


\bibitem[Yan et~al\mbox{.}(2023)]%
        {yan2023criticalfl}
\bibfield{author}{\bibinfo{person}{Gang Yan}, \bibinfo{person}{Hao Wang}, \bibinfo{person}{Xu Yuan}, {and} \bibinfo{person}{Jian Li}.} \bibinfo{year}{2023}\natexlab{}.
\newblock \showarticletitle{CriticalFL: A Critical Learning Periods Augmented Client Selection Framework for Efficient Federated Learning}. In \bibinfo{booktitle}{\emph{Proceedings of the 29th ACM SIGKDD Conference on Knowledge Discovery and Data Mining}}. \bibinfo{pages}{2898--2907}.
\newblock


\bibitem[Yang et~al\mbox{.}(2019a)]%
        {Yang19}
\bibfield{author}{\bibinfo{person}{Wensi Yang}, \bibinfo{person}{Yuhang Zhang}, \bibinfo{person}{Kejiang Ye}, \bibinfo{person}{Li Li}, {and} \bibinfo{person}{Cheng-Zhong Xu}.} \bibinfo{year}{2019}\natexlab{a}.
\newblock \showarticletitle{Ffd: A federated learning based method for credit card fraud detection}. In \bibinfo{booktitle}{\emph{Big Data--BigData 2019: 8th International Congress, Held as Part of the Services Conference Federation, SCF 2019, San Diego, CA, USA, June 25--30, 2019, Proceedings 8}}. Springer, \bibinfo{pages}{18--32}.
\newblock


\bibitem[Yang et~al\mbox{.}(2019b)]%
        {yang2019ffd}
\bibfield{author}{\bibinfo{person}{Wensi Yang}, \bibinfo{person}{Yuhang Zhang}, \bibinfo{person}{Kejiang Ye}, \bibinfo{person}{Li Li}, {and} \bibinfo{person}{Cheng-Zhong Xu}.} \bibinfo{year}{2019}\natexlab{b}.
\newblock \showarticletitle{Ffd: A federated learning based method for credit card fraud detection}. In \bibinfo{booktitle}{\emph{Big Data--BigData 2019: 8th International Congress, Held as Part of the Services Conference Federation, SCF 2019, San Diego, CA, USA, June 25--30, 2019, Proceedings 8}}. Springer, \bibinfo{pages}{18--32}.
\newblock


\bibitem[Yang et~al\mbox{.}(2023)]%
        {yang2023}
\bibfield{author}{\bibinfo{person}{Xiyuan Yang}, \bibinfo{person}{Wenke Huang}, {and} \bibinfo{person}{Mang Ye}.} \bibinfo{year}{2023}\natexlab{}.
\newblock \showarticletitle{Dynamic Personalized Federated Learning with Adaptive Differential Privacy}. In \bibinfo{booktitle}{\emph{Thirty-seventh Conference on Neural Information Processing Systems}}.
\newblock


\bibitem[Yi et~al\mbox{.}(2023)]%
        {yi2023ua}
\bibfield{author}{\bibinfo{person}{Jingwei Yi}, \bibinfo{person}{Fangzhao Wu}, \bibinfo{person}{Bin Zhu}, \bibinfo{person}{Jing Yao}, \bibinfo{person}{Zhulin Tao}, \bibinfo{person}{Guangzhong Sun}, {and} \bibinfo{person}{Xing Xie}.} \bibinfo{year}{2023}\natexlab{}.
\newblock \showarticletitle{UA-FedRec: untargeted attack on federated news recommendation}. In \bibinfo{booktitle}{\emph{Proceedings of the 29th ACM SIGKDD Conference on Knowledge Discovery and Data Mining}}. \bibinfo{pages}{5428--5438}.
\newblock


\bibitem[Yin et~al\mbox{.}(2021)]%
        {Yin21}
\bibfield{author}{\bibinfo{person}{Hongxu Yin}, \bibinfo{person}{Arun Mallya}, \bibinfo{person}{Arash Vahdat}, \bibinfo{person}{Jose~M Alvarez}, \bibinfo{person}{Jan Kautz}, {and} \bibinfo{person}{Pavlo Molchanov}.} \bibinfo{year}{2021}\natexlab{}.
\newblock \showarticletitle{See through gradients: Image batch recovery via gradinversion}. In \bibinfo{booktitle}{\emph{Proceedings of the IEEE/CVF Conference on Computer Vision and Pattern Recognition}}. \bibinfo{pages}{16337--16346}.
\newblock


\bibitem[Zhang et~al\mbox{.}(2020a)]%
        {Zhang20}
\bibfield{author}{\bibinfo{person}{Chengliang Zhang}, \bibinfo{person}{Suyi Li}, \bibinfo{person}{Junzhe Xia}, \bibinfo{person}{Wei Wang}, \bibinfo{person}{Feng Yan}, {and} \bibinfo{person}{Yang Liu}.} \bibinfo{year}{2020}\natexlab{a}.
\newblock \showarticletitle{BatchCrypt: Efficient homomorphic encryption for Cross-Silo federated learning}. In \bibinfo{booktitle}{\emph{2020 USENIX annual technical conference (USENIX ATC 20)}}. \bibinfo{pages}{493--506}.
\newblock


\bibitem[Zhang et~al\mbox{.}(2020b)]%
        {zhang2020gan}
\bibfield{author}{\bibinfo{person}{Jingwen Zhang}, \bibinfo{person}{Jiale Zhang}, \bibinfo{person}{Junjun Chen}, {and} \bibinfo{person}{Shui Yu}.} \bibinfo{year}{2020}\natexlab{b}.
\newblock \showarticletitle{Gan enhanced membership inference: A passive local attack in federated learning}. In \bibinfo{booktitle}{\emph{ICC 2020-2020 IEEE International Conference on Communications (ICC)}}. IEEE, \bibinfo{pages}{1--6}.
\newblock


\bibitem[Zhao et~al\mbox{.}(2020)]%
        {Zhao20}
\bibfield{author}{\bibinfo{person}{Bo Zhao}, \bibinfo{person}{Konda~Reddy Mopuri}, {and} \bibinfo{person}{Hakan Bilen}.} \bibinfo{year}{2020}\natexlab{}.
\newblock \showarticletitle{idlg: Improved deep leakage from gradients}.
\newblock \bibinfo{journal}{\emph{arXiv preprint arXiv:2001.02610}} (\bibinfo{year}{2020}).
\newblock


\bibitem[Zhao et~al\mbox{.}(2019)]%
        {zhao2019differential}
\bibfield{author}{\bibinfo{person}{Jingwen Zhao}, \bibinfo{person}{Yunfang Chen}, {and} \bibinfo{person}{Wei Zhang}.} \bibinfo{year}{2019}\natexlab{}.
\newblock \showarticletitle{Differential privacy preservation in deep learning: Challenges, opportunities and solutions}.
\newblock \bibinfo{journal}{\emph{IEEE Access}}  \bibinfo{volume}{7} (\bibinfo{year}{2019}), \bibinfo{pages}{48901--48911}.
\newblock


\bibitem[Zheng et~al\mbox{.}(2021)]%
        {zheng2021federated}
\bibfield{author}{\bibinfo{person}{Wenbo Zheng}, \bibinfo{person}{Lan Yan}, \bibinfo{person}{Chao Gou}, {and} \bibinfo{person}{Fei-Yue Wang}.} \bibinfo{year}{2021}\natexlab{}.
\newblock \showarticletitle{Federated meta-learning for fraudulent credit card detection}. In \bibinfo{booktitle}{\emph{Proceedings of the Twenty-Ninth International Conference on International Joint Conferences on Artificial Intelligence}}. \bibinfo{pages}{4654--4660}.
\newblock


\bibitem[Zhu et~al\mbox{.}(2019)]%
        {Zhu19}
\bibfield{author}{\bibinfo{person}{Ligeng Zhu}, \bibinfo{person}{Zhijian Liu}, {and} \bibinfo{person}{Song Han}.} \bibinfo{year}{2019}\natexlab{}.
\newblock \showarticletitle{Deep leakage from gradients}.
\newblock \bibinfo{journal}{\emph{Advances in neural information processing systems}}  \bibinfo{volume}{32} (\bibinfo{year}{2019}).
\newblock


\end{thebibliography}

\end{document}


\title{FedRE: \underline{R}obust and \underline{E}ffective Federated Learning \\with Privacy Preference (Supplementary Material)}

\maketitle

\appendix

\section{Experiment Settings}
\label{app-experiment settings}
For the \textbf{T-SROIE} dataset, we use a combination of manual annotation and the OCR labeled bbox in the source dataset to jointly determine the sensitive regions. The manually labeled sensitive regions of the receipts image are mainly defined as consumption tax number, invoice number, amount total and other regional content to ensure that the overall labeling content and labeling information is more consistent; in addition,the sensitive regions of the same format bills to maintain uniformity to ensure the consistency of the labeling logic before and after. Then we intersect the manually labeled sensitive region with the bbox of the OCR labeling, if the intersection ratio of the region of the manually labeled region and the region of an OCR labeled bbox is greater than 20\%, the OCR labeled bbox is identified as a sensitive region.

For the \textbf{DocTamper} dataset, the privacy regions in contracts, invoices, and receipts can be categorized into image privacy sensitive regions such as official seals, and text privacy sensitive regions such as document numbers and amounts.We use manual annotation to determine the image privacy regions and OCR recognition to annotate the text privacy sensitive regions, respectively. According to the practical experience, we set the upper limit of the text privacy region of a single image as 8, and select the number of private text boxes of an image according to a certain distribution probability, and then determine a number of text boxes as the text privacy sensitive region of the image based on the number of text boxes obtained and the results of regular matching of the text content in the text boxes.

The \textbf{public dataset} on the central server used in our experiments is split from the two training datasets, and we select 100,000 sampls in the DocTamper training set as the client's training data, and 200 data in the remaining data as the public dataset on the central server. Each sample in these public data belongs to one of the four formats of contract images, invoices, normal pages, receipts and notes, like the training data on clients. The central server averages the PSI computations for each format (50 samples per format) as the basis for estimating the  distribution of noise in gradients that trained from client's praivate data in same format. Because the format of the public dataset on the central server covers the common data formats in client's training set. The public dataset contains data in the same format as the private dataset but with different content, and the PSI scores for data in the same format are similar, so a public dataset is enough to estimate PSI scores of the client dataset. Similarly, for the invoice data in the four formats in the T-SROIE training set , we selected 600 of these samples as clients' training data and 20 samples (5 samples per format) as public dataset on the central server. 

For the experiments with 1, 2, and 5 clients, we train 10 communication rounds, for the experiments with 10 clients, we train 20 communication rounds, and for the experiments with 20 clients, we train 30 communication rounds, the learning rate of all experiments is chosen as 0.005, the training set is selected as described before, the number of private data of each client is the same in training, but the format distribution is different from that of all other clients.

\section{Evaluation Metrics}
\label{app-evaluation metrics}
In line with the methodologies outlined in ~\cite{Wang22} and ~\cite{Li23}, we rigorously evaluate the protection effectiveness of our proposed method by analyzing the similarity between the reconstructed and original images specifically within privacy-sensitive regions. To achieve this, we follow those methods and employ four key metrics that offer complementary insights into the quality and fidelity of the reconstructed images.  

\textbf{Mean Square Error (MSE)} quantifies the average squared difference between corresponding pixel values in the privacy-sensitive regions of the original and reconstructed images. This metric serves as a basic indicator of the level of distortion introduced during the protection process, with higher MSE values indicating lower similarity, and thus better protection performance.  

\textbf{Peak Signal-to-Noise Ratio (PSNR)} is a widely used metric in image quality assessment. It provides a logarithmic measure of the ratio between the maximum possible power of a signal and the power of corrupting noise that affects the fidelity of its representation. In our context, a lower PSNR value suggests that the reconstructed image in the privacy-sensitive regions is not similar to the original, indicating better protection while maintaining visual quality.  

\textbf{Structural Similarity Index Metric (SSIM)} assesses the perceived quality of the reconstructed image by considering structural information in addition to luminance and contrast. SSIM values range from -1 to 1, with 1 indicating perfect structural similarity. This metric is useful in evaluating how well the method preserves the visual structures within privacy-sensitive regions. A lower SSIM indicates better protection.

\textbf{Perceptual Similarity Score (LPIPS)}, a learned perceptual metric that captures the differences in human perception between images. By leveraging deep neural network features, LPIPS provides a more nuanced assessment of the perceptual similarity between the original and reconstructed images in privacy-sensitive regions. This metric is valuable in evaluating how well the method preserves the visual fidelity that is meaningful to human observers, with a higher LPIPS indicating better protection.

It is important to note that all four similarity metrics—MSE, PSNR, SSIM, and LPIPS—are focused exclusively on the privacy-sensitive regions of the image, ensuring that our evaluation accurately reflects the performance of the method in protecting these critical areas.  

To further validate the effectiveness of our approach, we also evaluate the training performance of the tamper detection task using Intersection over Union (\textbf{IoU}), \textbf{Precision}, \textbf{Recall}, and \textbf{F-score}. IoU measures the overlap between the predicted and ground truth tampering regions, providing a spatial accuracy assessment. Precision quantifies the fraction of correctly predicted tampering pixels among all predicted tampering pixels, while Recall measures the fraction of correctly predicted tampering pixels among all actual tampering pixels. Finally, F-score is the harmonic mean of Precision and Recall, offering a balanced assessment of the tamper detection system's performance. These metrics collectively provide a comprehensive evaluation of whether the tamper detection can accurately identify tampering while minimizing false positives and negatives.

\bibliographystyle{ACM-Reference-Format}
\bibliography{sample-base}